\documentclass{article}

\PassOptionsToPackage{round,authoryear}{natbib}
 \usepackage[preprint]{neurips_2025}

\usepackage[utf8]{inputenc} 
\usepackage[T1]{fontenc}    
\usepackage{hyperref}       
\usepackage{url}            
\usepackage{booktabs}       
\usepackage{amsfonts}       
\usepackage{nicefrac}       
\usepackage{microtype}      
\usepackage{xcolor}         
\usepackage{xspace}

\usepackage[most]{tcolorbox}
\usepackage[table]{xcolor}

\usepackage[export]{adjustbox}

\usepackage{amssymb}
\usepackage{mathtools}

\usepackage{subcaption}
\usepackage{graphicx}
\usepackage{etoolbox}

\usepackage{wrapfig}

\usepackage{booktabs}
\usepackage{multirow}
\usepackage{amsmath}
\usepackage{makecell}

\usepackage{algorithm}
\usepackage{algorithmic}

\usepackage[most]{tcolorbox}
\usepackage{listings}
\usepackage{xcolor}

\definecolor{boxbg}{gray}{0.96}
\definecolor{boxframe}{gray}{0.35}

\lstdefinestyle{mytt}{
  basicstyle=\ttfamily\footnotesize,
  columns=fullflexible,
  breaklines=true,
  showstringspaces=false,
  numbers=left,
  numberstyle=\tiny\color{gray},
  numbersep=6pt,
  xleftmargin=1.2em,
  frame=none
}

\newtcolorbox{promptbox}[1][]{
  enhanced,
  breakable,              
  colback=boxbg,
  colframe=boxframe,
  boxrule=0.8pt,
  arc=2mm,                
  left=2mm,right=2mm,top=1.5mm,bottom=1.5mm,
  #1
}

\definecolor{envbg}{RGB}{255, 247, 237}   
\definecolor{envframe}{RGB}{230, 140, 60} 
\newtcolorbox{envbox}[1][]{
  enhanced,
  breakable,              
  colback=envbg,
  colframe=envframe,
  boxrule=0.8pt,
  arc=2mm,                
  left=2mm,right=2mm,top=1.5mm,bottom=1.5mm,
  #1
}

\definecolor{herabg}{RGB}{240, 247, 255}
\definecolor{heraframe}{RGB}{100, 150, 220}
\newtcolorbox{herabox}[1][]{
  enhanced,
  breakable,              
  colback=herabg,
  colframe=heraframe,
  boxrule=0.8pt,
  arc=2mm,                
  left=2mm,right=2mm,top=1.5mm,bottom=1.5mm,
  #1
}

\definecolor{devicebg}{RGB}{240, 252, 245}   
\definecolor{deviceframe}{RGB}{110, 170, 140} 
\newtcolorbox{devicebox}[1][]{
  enhanced,
  breakable,
  colback=devicebg,
  colframe=deviceframe,
  boxrule=0.8pt,
  arc=2mm,
  left=2mm,right=2mm,top=1.5mm,bottom=1.5mm,
  #1
}

\definecolor{cloudbg}{RGB}{248, 245, 255}    
\definecolor{cloudframe}{RGB}{150, 130, 200} 
\newtcolorbox{cloudbox}[1][]{
  enhanced,
  breakable,
  colback=cloudbg,
  colframe=cloudframe,
  boxrule=0.8pt,
  arc=2mm,
  left=2mm,right=2mm,top=1.5mm,bottom=1.5mm,
  #1
}

\ifodd 1
\definecolor{darkgreen}{RGB}{0,200,0}
\else

\fi

\newcommand{\name}{Hera\xspace} 

\title{\raisebox{-0.2\height}{\includegraphics[height=1.2em]{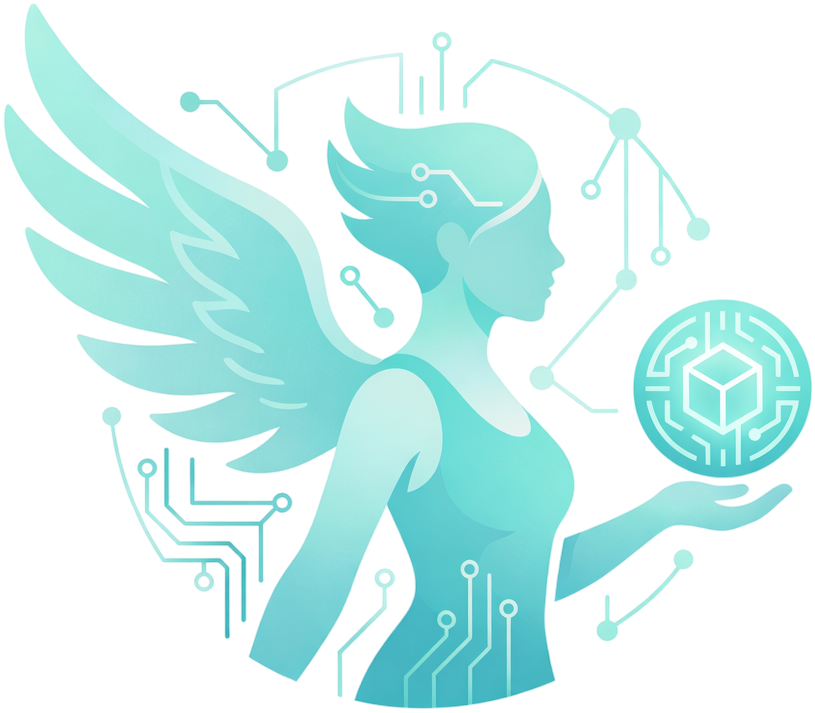}}\xspace \name: Learning Long-Horizon Coordination for Device--Cloud Collaborative LLM Agents}

\author{
Yuxin Zhang$^{1,2}$ \\ \And
Mengxue Hu$^{1}$ \\ \And
Zheng Lin$^{3}$ \\ \And
Xiaoyi Fan$^{4}$ \\ \And
Fan Xie$^{5}$ \\ \And
Zihan Fang$^{1}$ \\ \And
Jing Yang$^{6}$ \\ \And
Wenjun Zhu$^{1,7}$ \\ \And
Zhiwen Chen$^{2}$ \\ \And
Chengfei Lv$^{2}$ \\ \And
Zhe Chen$^{1,7,*}$\\ \And
\texttt{$^{1}$Fudan University \quad
$^{2}$Alibaba Group \quad $^{3}$The University of Hong Kong}\\
\texttt{$^{4}$Shenzhen MSU-BIT University \quad  $^{5}$New York University \quad $^{6}$Universiti Malaya} \\\texttt{$^{7}$SpaceAIC Co., Ltd.}
}

\begin{document}

\maketitle

\begin{abstract}
Large language model (LLM) agents excel at solving complex long-horizon tasks through autonomous interaction with environments. However, their real-world deployment faces a fundamental \textit{device--cloud dilemma}: 
on-device models are efficient but often brittle, while cloud models are stronger but costly in computation. State-of-the-art LLM device--cloud routers usually make coarse task-level decisions, which cannot adapt to the changing difficulty of multi-step agent interactions.
To address this issue, we present \name, a \textit{step-level} device--cloud LLM agent coordinator for long-horizon tasks achieving a strong performance--cost Pareto frontier.
\name adopts a novel two-stage training paradigm: (1) imitation learning for cold-start, followed by (2) reinforcement learning that jointly optimizes task success and cloud usage efficiency.
The first stage casts step-level routing as a supervised classification problem: the device agent is replayed on cloud trajectories, with each state labeled by the agreement between device and cloud actions.
In the second stage, we perform cost-aware reinforcement learning by grouping identical states across trajectories and updating \name with labels favoring higher expected return and fewer future cloud calls.
We evaluate \name on ALFWorld, WebShop, and AppWorld, where it consistently outperforms prior methods,
achieving 92.5\% of the cloud-only success rate with cloud use in only 46.3\% of steps.
\end{abstract}



\section{Introduction}
Recent advances in large language models (LLMs)~\citep{singh2025openai,yang2025qwen3,liu2024deepseek}, combined with agentic frameworks such as ReAct~\citep{Yao2023ReAct}, have empowered LLM agents to perform complex, long-horizon reasoning and decision-making well beyond static problem solving~\citep{chang2024agentboard}.
These agents operate in open-ended environments through multi-step interaction loops, using step-wise feedback to guide subsequent actions, and have demonstrated strong performance across diverse domains, including software engineering~\citep{zhang2024codeagent,yang2025kimi}, GUI navigation~\citep{vattikondatrain,gurreal}, interactive gameplay~\citep{wangvoyager,tan2025lumine}, and robotic control~\citep{zitkovich2023rt,hanrobocerebra}.

Despite their promise, LLM agents in real-world environments face a fundamental device--cloud trade-off.
On one hand, deploying advanced agents directly on user-end is infeasible due to stringent resource constraints.
Small models (e.g., $<$10B parameters) that fit within device memory limits typically exhibit significant performance degradation in complex reasoning tasks, often failing to maintain long-term coherence in multi-step trajectory planning~\citep{nozaki2025efficient,ong2025routellm}. On the other hand, reliance on powerful cloud-based closed-source models introduces substantial network latency and high usage cost, limiting their practicality for real-time applications such as robotic manipulation and interactive GUI navigation~\citep{ravichandran2025distilling,wanyanlook}.
For example, a cloud-based drone navigation agent may miss dynamic obstacles due to network jitter, whereas an on-device model may lack the “System 2” reasoning depth to plan an optimal global path.

To balance local efficiency and global performance, recent work dynamically allocates inference between device- and cloud-side LLMs~\citep{narayan2025cost,li2025collaborative}.
Existing methods treat each request as an atomic unit, making a one-shot routing decision before inference begins by assigning the entire query to a single model based on estimated difficulty, confidence, or preference~\citep{ong2025routellm,Eagle2024}. While effective for standalone tasks such as question answering, summarization, and code completion, this paradigm is poorly suited to long-horizon agent settings, where execution unfolds as a sequence of interdependent steps whose difficulty changes with the evolving environment state~\citep{fenggroup,dong2026agentic}.



\begin{figure*}[t]
    \centering
    \vspace{-1.0em}  
\includegraphics[width=\textwidth,height=8.9cm,keepaspectratio]{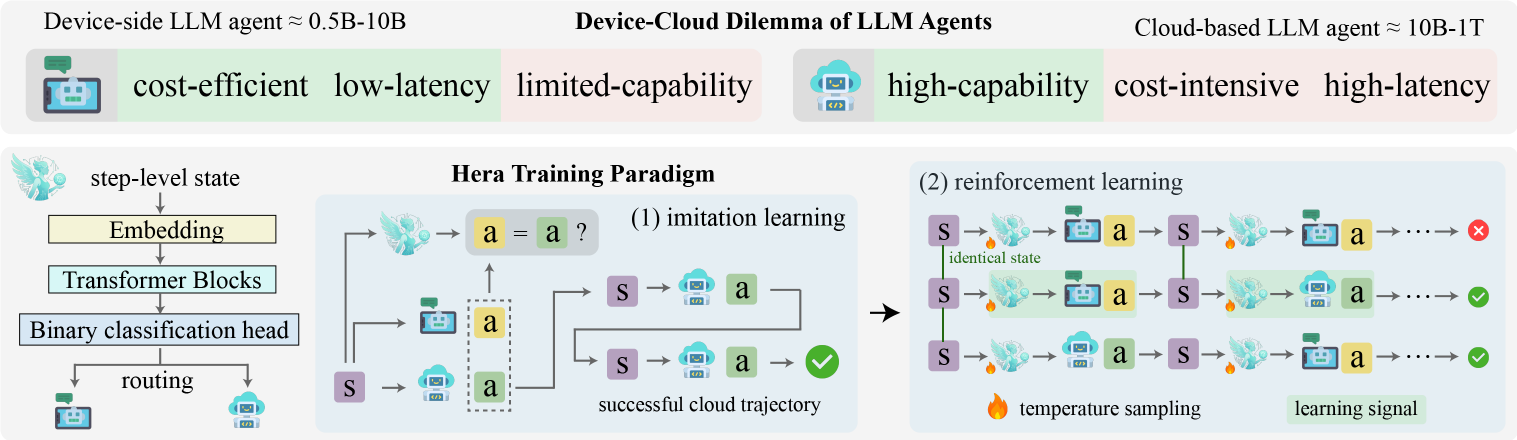}
\caption{LLM agents face a fundamental device--cloud dilemma. To address this, \name is proposed as a lightweight step-level device--cloud coordinator trained via a two-stage paradigm: imitation learning for cold-start, followed by reinforcement learning for performance--cost co-optimization.}
    \vspace{-1.5em}  
\label{figure:teaser}
\end{figure*}

To address this limitation, this paper investigates \textit{step-level} device--cloud routing for long-horizon LLM agent tasks, aiming to better balance task success and inference cost through routing decisions made at each environment interaction rather than once per task. Importantly, we find that offloading only a small fraction of critical steps along the trajectory to the cloud is sufficient to match the task success of cloud-only execution (\S\ref{sec:motivation}).
Building on this insight, we propose \name, a lightweight yet powerful device--cloud coordinator that identifies when cloud assistance is necessary, reducing cloud usage while maintaining strong task completion.
\name follows a two-stage training paradigm, as shown in Figure~\ref{figure:teaser}: (1) imitation learning for routing cold-start (\S\ref{sec:hera_il}), and (2) reinforcement learning for cost-performance co-optimization (\S\ref{sec:hera_rl}).
We initialize \name via imitation learning by replaying the device agent on cloud trajectories and labeling each step by device--cloud action consistency.
We then refine \name with cost-aware reinforcement learning: across multiple rollouts, we group identical states and assign step-level preference labels that prioritize higher estimated return and lower expected future cloud usage.
Across various challenging agent benchmarks including ALFWorld~\citep{ALFWorld20}, WebShop~\citep{yao2022webshop}, and AppWorld~\citep{appworld}, \name consistently outperforms existing methods, demonstrating superior success--cost co-optimization.

Our contributions are threefold:
\begin{itemize}
    \item To the best of out knowledge, \name is the first step-level device--cloud agent coordinator, addressing the performance--cost trade-off of deploying LLM agents in long-horizon tasks.
    \item We develop a two-stage training framework for \name: an imitation-learning stage that initializes routing from device--cloud action agreement, followed by a state-grouped reinforcement learning stage that jointly optimizes trajectory return and cloud cost.
    \item Experiments on various agent benchmarks show that \name outperforms state-of-the-art methods, achieving a stronger trade-off between task success and cloud usage.
\end{itemize}

\section{Related work}
\label{subsec::related_work}

\paragraph{LLMs agents.}
Through carefully designed prompting~\citep{Yao2023ReAct,shinn2023reflexion}, enhanced memory and retrieval mechanisms~\citep{wang2024mobile,xu2025amem}, and integration with external tools~\citep{schick2023toolformer,xie2024osworld}, LM agents are capable of executing complex, multi-step tasks that were previously beyond reach~\citep{chang2024agentboard,xie2024travelplanner,wangvagen}.
Recent work uses supervised fine-tuning~\citep{xia2025sand,zhang2024you,li2026reasoning} and, in particular, reinforcement learning~\citep{fenggroup,mai2025agentic,zhang2026land} to enable LLM agents to learn directly from environment interactions rather than static prompts or handcrafted workflows, significantly enhancing long-horizon reasoning and multi-step decision-making and forming a promising paradigm for general-purpose autonomous intelligence~\citep{comanici2025gemini,qwen3.5,liu2025deepseek}.
However, LLM agents face a fundamental device--cloud dilemma: user-end devices are constrained in memory and energy, limiting LLM scale and performance, while offloading to cloud models adds latency, hindering latency-sensitive applications such as robotics and mobile assistants, and driving up operational costs.

\paragraph{Device--cloud LLM collaboration.}
Recent work has explored device--cloud LLM routing, which selectively dispatches each request to models of different capacities to trade off output quality against latency and cost.
FrugalGPT~\citep{chen2024frugalgpt} and Hybrid LLM~\citep{ding2024hybrid} route requests based on estimated quality--cost trade-offs, while Eagle~\citep{Eagle2024} and RouteLLM~\citep{ongroutellm} use historical similarity or preference-based win prediction to select the stronger model under a budget.
Another line of work explores tighter device--cloud collaboration, where the cloud decomposes or verifies subtasks and the device executes them over local context to reduce cloud cost.
For example, MinionS~\citep{narayan2025cost} lets the cloud decompose or verify subtasks while the device executes them over local context to reduce cloud cost.
However, these methods rely on request-level routing or coarse subtask decomposition, making them ill suited to long-horizon agent tasks, where evolving environments and step-varying difficulty demand finer-grained coordination.

\section{Preliminaries}
\label{sec:preliminary}

\paragraph{Problem setup.}
We consider a finite-horizon, multi-step agentic task setting in which an LLM-parameterized policy $\pi_\theta$ governs an agent interacting with an environment to accomplish a task specified by a description $x \in \mathcal{X}$.
At each step \(t \in \{1,\dots,T\}\), the agent observes \(\mathbf{s}_t \in \mathcal{S}\) and samples a textual action \(\mathbf{a}_t \in \mathcal{V}^{n}\) according to the conditional distribution \(\pi_\theta(\mathbf{a}_t \mid \mathbf{s}_t , x)\), where $\mathcal{V}$ is the LLM vocabulary and $n$ is the maximum output length.
The environment subsequently returns a scalar reward $r_t \in \mathbb{R}$ and the next state $\mathbf{s}_{t+1}$.
An episode produces a trajectory \(\tau = \{(\mathbf{s}_1, \mathbf{a}_1 ,r_1), (\mathbf{s}_2, \mathbf{a}_2 ,r_2),\dots, (\mathbf{s}_T, \mathbf{a}_T ,r_T)\}\) with an associated return \(R(\tau)\) (e.g., verifiable, outcome-level task feedback), which is often used to optimize the LLM policy to maximize the expected return of the agent’s trajectories~\citep{schulman2017ppo,shao2024deepseekmath}.

\paragraph{Device--cloud collaboration.}
Extending the single-agent paradigm, we formalize the device--cloud collaborative system: the device policy $\pi_{\theta_\mathrm{d}}$ performs low-latency inference using a lightweight LLM (typically $<$3B parameters) under strict memory and compute constraints, while the cloud policy $\pi_{\theta_\mathrm{c}}$ exploits massive scale (hundreds of billions of parameters) to deliver superior reasoning capabilities at the cost of non-negligible latency and high resource overhead.
Let $r_\phi : \mathcal{S} \times \mathcal{X} \rightarrow \{0,1\}$ denote a device--cloud router parameterized by $\phi$, where for a task--state pair $(\mathbf{s},x)$, $r_\phi(\mathbf{s},x)=0$ selects local processing by $\pi_{\theta_\mathrm{d}}$, and $r_\phi(\mathbf{s},x)=1$ escalates the request to $\pi_{\theta_\mathrm{c}}$.
Thus, the final action of device--cloud collaborative agents under $(\mathbf{s},x)$ is given by:
\begin{equation}
\hat{\mathbf{a}}(\mathbf{s},x)
=
\big(1 - r_\phi(\mathbf{s},x)\big)\,\pi_{\theta_\mathrm{d}}(\mathbf{s},x)
+
r_\phi(\mathbf{s},x)\,\pi_{\theta_\mathrm{c}}(\mathbf{s},x).
\end{equation}

\paragraph{Optimization objective.}
The objective of \name is to maximize operational efficiency by minimizing cloud offloading, while maintaining task performance competitive with a cloud-only baseline.
We formulate the optimization of device--cloud router $\phi$ as follows:
\begin{equation}
\begin{aligned}
\min_{\phi} \quad & 
\mathbb{E}_{x \sim p(X), \, \tau \sim \pi_{\theta_\mathrm{mix}}}
\Bigg[ \sum_{t=1}^T r_\phi(\mathbf{s}_t, x) \Bigg] \\
\text{s.t.} \quad &
\mathbb{E}_{x \sim p(X), \, \tau \sim \pi_{\theta_\mathrm{mix}}}
\big[ R(\tau) \big] 
\ge
\mathbb{E}_{x \sim p(X), \, \tau \sim \pi_{\theta_\mathrm{c}}} 
\big[ R(\tau) \big].
\end{aligned}
\label{eq:opt_obj}
\end{equation}
where $\pi_{\theta_\mathrm{mix}} = \big(1 - r_\phi(\mathbf{s},x)\big)\,\pi_{\theta_\mathrm{d}}+r_\phi(\mathbf{s},x)\,\pi_{\theta_\mathrm{c}}$ is the collaborative policy controlled by $\phi$.

\section{Do We Need the Cloud LLM for Every Step?}
\label{sec:motivation}

This section presents an in-depth investigation of the performance gap between device and cloud LLM agents on long-horizon tasks.
We use Qwen2.5-7B-Instruct~\citep{yang2024qwen25} as the device model and its closed-source counterpart Qwen-Max~\citep{qwen2.5-max2025} as the cloud model, representing a resource-efficient model for local deployment and a compute-intensive, state-of-the-art oracle, respectively.
We evaluate them on two demanding long-horizon agent benchmarks: ALFWorld~\citep{ALFWorld20}, which focuses on embodied household task completion, and WebShop~\citep{yao2022webshop}, which requires multi-step e-commerce navigation and decision-making.
More experimental details are shown in \S~\ref{subsec:exp_set} and Appendix~\ref{appendix:exp}.

\begin{figure}[t]
    \centering
    \begin{subfigure}[t]{0.234\textwidth}
        \centering
        \includegraphics[width=\textwidth, valign=t]{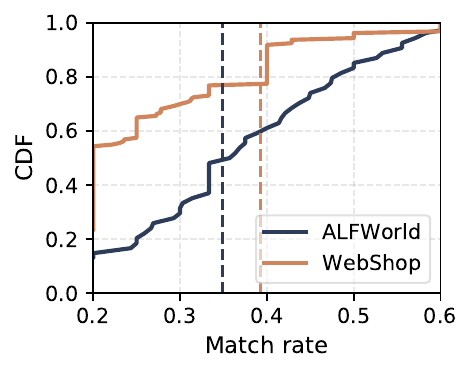}
        \vspace{-0.55em}
        \caption{Step-level match.}
        \label{subfig:edge_match_cdf}
    \end{subfigure}
    \begin{subfigure}[t]{0.234\textwidth}
        \centering
        \includegraphics[width=\textwidth, valign=t]{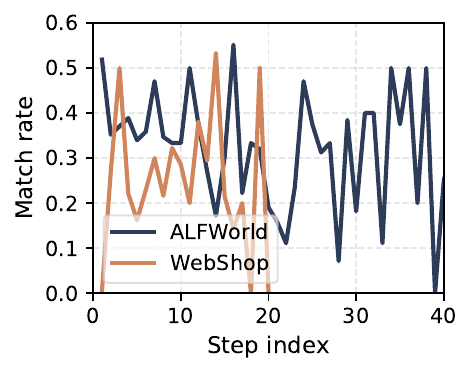}
        \vspace{-0.55em}
        \caption{Step position.}
        \label{subfig:step_pos}
    \end{subfigure}
    \begin{subfigure}[t]{0.234\textwidth}
        \centering
        \includegraphics[width=\textwidth, valign=t]{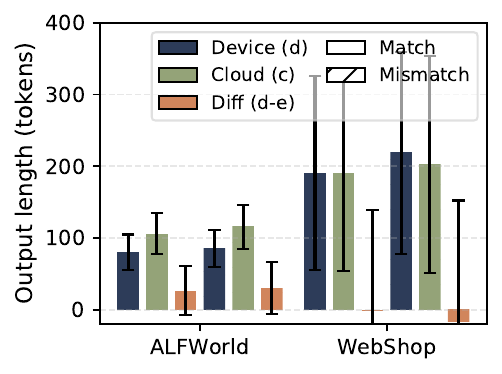}
        \caption{Reasoning length.}
        \label{subfig:moti_thinking_length}
    \end{subfigure}
    \begin{subfigure}[t]{0.234\textwidth}
        \centering
        \includegraphics[width=\textwidth, valign=t]{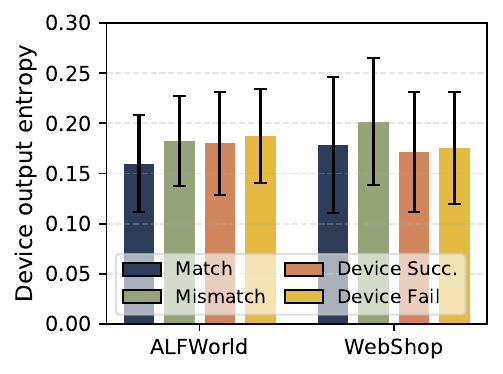}
        \caption{Predictive entropy.}
        \label{subfig:moti_ent}
    \end{subfigure}\\
    \begin{subfigure}[t]{0.99\textwidth}
        \centering
        \includegraphics[width=\textwidth, valign=t]{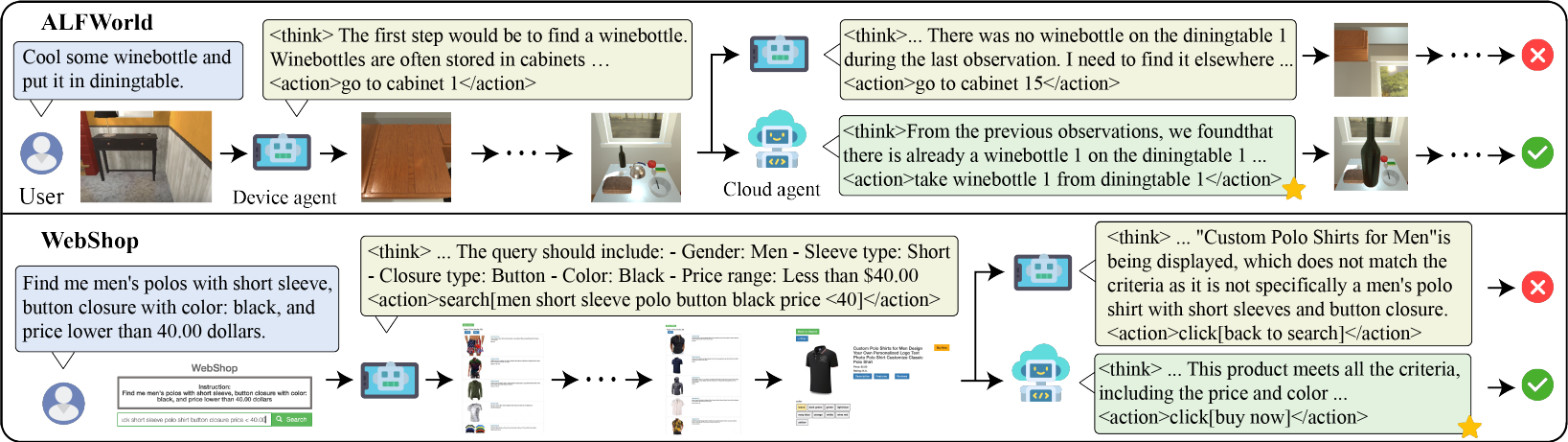}
        \caption{Example trajectories of device--cloud agents collaboratively solving long-horizon tasks.}
        \label{fig:moti_traj_example}
    \end{subfigure}
    \vspace{-.5em}
    \caption{Step-level performance comparison of device Qwen2.5-7B-Instruct and cloud Qwen-Max on ALFWorld and WebShop tasks.}
    \label{fig:moti_exp_stepwise}
    \vspace{-1.5em}
\end{figure}

Individually, device and cloud LLMs achieve success rates of 48.6\% (3.8\%) and 90.9\% (35.6\%) on ALFWorld (WebShop), respectively, revealing a substantial \textit{task-level} performance gap.
However, not all steps in device-failed trajectories require cloud intervention.
As shown in Figure~\ref{subfig:edge_match_cdf}, for tasks where the device fails but the cloud succeeds, we replay the cloud trajectory on the device by providing the same state at each step, and find that 34.9\% (39.3\%) of steps produce identical outputs on ALFWorld (WebShop).
\textit{This suggests that an ideal step-level device--cloud agent router would need to invoke the cloud on $<$25\% of steps to match cloud-only task success.}
We further analyze device--cloud step-level agreement across multiple dimensions, including trajectory position, reasoning length, and predictive entropy.
Figure~\ref{subfig:step_pos} shows the step match rate across trajectory positions (up to 50 steps for ALFWorld and 20 for WebShop).
Notably, the match rate varies with step index rather than remaining stable, reflecting heterogeneous task structure and stage-dependent reasoning difficulty.
For instance, in ALFWorld, early steps primarily involve environment exploration, while intermediate stages require more complex planning and long-horizon reasoning, leading to increased device--cloud divergence. Case studies are provided later to further illustrate these patterns.

Figures~\ref{subfig:moti_thinking_length} and~\ref{subfig:moti_ent} quantify the relationship between device--cloud step-level agreement and action difficulty from the perspectives of reasoning length and predictive entropy, respectively.
Reasoning length reflects the depth of deliberation required at each step, while sequence-level predictive entropy captures model uncertainty and is computed as the average token-level entropy:
\begin{equation}
H(\mathbf{a}_t) = -\frac{1}{n} \sum_{i=1}^{n} \sum_{v \in \mathcal{V}} p_\theta(v \mid y_{<i}, \mathbf{s}_t, x) \log p_\theta(v \mid y_{<i}, \mathbf{s}_t, x),
\end{equation}
where $\mathbf{a}_t = (y_1, \dots, y_n) \in \mathcal{V}^n$ is the generated action sequence at step $t$, and $y_{<i} = (y_1, \dots, y_{i-1})$ denotes the token prefix.
The results show that steps with consistent device--cloud outputs are, on average, 13.8 tokens shorter than mismatched ones, indicating a lower required reasoning depth (i.e., more direct and simpler).
Interestingly, the device--cloud reasoning-length gap differs markedly across the two benchmarks.
On ALFWorld, the structured environment and limited action space allow the device model to act more directly, though often without capturing deeper dependencies handled by the cloud. In contrast, WebShop presents a large, unstructured action space—e.g., free-form queries and diverse attribute parsing—causing the weaker device model to produce more verbose deliberation, while the stronger cloud model remains more concise and efficient.
Beyond reasoning length, entropy also exhibits a clear separation: the device model is more uncertain on mismatched steps and more confident on those aligned with the cloud, suggesting that the need for cloud intervention is structured by step difficulty rather than random chance.
Moreover, entropy exhibits a similar benchmark-dependent pattern: cloud interventions reduce device entropy on ALFWorld, indicating stronger alignment with device reasoning, but increase it on WebShop, where cloud decisions more often fall outside the device model’s familiar decision regime.

Figure~\ref{fig:moti_traj_example} shows examples of device--cloud step-level agreement, with more cases and analysis provided in Appendix~\ref{appendix:case}.
While the device agent can handle many routine follow-up actions, its gap from the cloud agent is concentrated in decisions requiring deeper context tracking, precise grounding, or complex constraint composition. Accurately identifying these steps could substantially reduce cloud calls while preserving task success, highlighting the promise of step-level coordination.

\section{\name: A Long-Horizon Device--Cloud LLM Agent Coordinator}
\label{sec:hera}

This section presents \name, a step-level device--cloud agent coordinator that optimizes Eq.~\ref{eq:opt_obj} to maximize task performance while minimizing cloud reliance.
\name consists of a 500M-parameter decoder-only Transformer with a binary classification head to route between device agent $\pi_{\theta_\mathrm{d}}$ and cloud agent $\pi_{\theta_\mathrm{c}}$. The training of \name follows a two-stage pipeline (pseudocode in Appendix~\ref{append:code}): \textit{a cold-start phase with imitation learning}, followed by \textit{a refinement phase with reinforcement learning}, ultimately enabling effective performance--cost co-optimization in device--cloud LLM agent systems.

\subsection{Cold Start with Imitation Learning}
\label{sec:hera_il}
To equip \name (the coordinator $r_\phi$ in \S~\ref{sec:preliminary}) with initial routing capability, we first employ a cold-start imitation learning (IL) phase. The objective in this stage is to transform the discrete routing decision into a supervised classification problem by identifying the capability gap between device and cloud models across long-horizon trajectories.
For each task $x$, we begin by rolling out both device-side $\pi_{\theta_\mathrm{d}}$ and cloud-side $\pi_{\theta_\mathrm{c}}$ to collect trajectories $\tau_{\mathrm{d}} (x)$ and $\tau_{\mathrm{c}}(x)$, respectively.
Tasks exhibiting the pronounced device--cloud performance gaps are then selected as:
\begin{equation}
X_{\text{diff}} = \left\{ x \;\middle|\; R(\tau_{\mathrm{c}}(x)) - R(\tau_{\mathrm{d}}(x)) > \delta \right\},
\end{equation}
where $\delta$ serves as a prescribed margin of sub-optimality. This set delineates a regime where cloud assistance is fundamentally necessary.

Using $X_{\text{diff}}$, we construct step-level supervision signals for $r_\phi$ by using cloud trajectories as an oracle reference, which provide guidance on states where $\pi_{\theta_\mathrm{d}}$ and $\pi_{\theta_\mathrm{c}}$ exhibit discrepancies. 
Specifically, for each $x \in X_{\text{diff}}$, we take cloud rollout 
$\tau_{\mathrm{c}}(x) = \{({\mathbf{s}}^{\mathrm{c}}_t, {\mathbf{a}}_t^{\mathrm{c}})\}_{t=1}^T$, and feed each $({\mathbf{s}}^\mathrm{c}_t,x)$ into $\pi_{\theta_\mathrm{d}}$ to get the corresponding action
$
{\mathbf{a}}_t^{\mathrm{e}} \sim \pi_{\theta_\mathrm{d}}(\cdot \mid {\mathbf{s}}^\mathrm{c}_t,x)
$.
We define a step-level consistency indicator as $
y_t = 1- \mathbb{I}\big[{\mathbf{a}}_t^{\mathrm{e}} = {\mathbf{a}}_t^{\mathrm{c}}\big]
$, where $y_t=1$ indicates that $\pi_{\theta_\mathrm{d}}$ output disagrees with the cloud oracle and current state should be routed to $\pi_{\theta_\mathrm{c}}$, while $y_t=0$ means $\pi_{\theta_\mathrm{d}}$ is sufficient for local execution.

Based on this, we construct a cold-start IL dataset for the $r_\phi$:
\begin{equation}
\mathcal{D}_{\mathrm{IL}} = \left\{ ({\mathbf{s}}_t, x, y_t) \;\middle|\; x \in X_{\text{diff}},\ ({\mathbf{s}}_t, {\mathbf{a}}_t^{\mathrm{c}}) \in \tau_{\mathrm{c}}(x) \right\}.
\end{equation}
The coordinator $r_\phi$ is then trained as a binary classifier on $\mathcal{D}_{\mathrm{IL}}$ with input $({\mathbf{s}}_t, x)$ and label $y_t$ (as defined in \S~\ref{sec:preliminary}) using a binary cross-entropy (BCE) loss:
\begin{equation}
\mathcal{L}_{\text{IL}}(\phi) = - \frac{1}{|\mathcal{D}_{\text{IL}}|} 
\sum_{(\mathbf{s}_t, x, y_t) \in \mathcal{D}_{\text{IL}}} 
\left[ 
y_t \log p_\phi(\mathbf{s}_t, x) 
+ (1 - y_t)\log \big(1 - p_\phi(\mathbf{s}_t, x)\big) 
\right],
\label{eq:il}
\end{equation}
where $p_\phi(\mathbf{s}, x) \in (0,1)$ denotes the Bernoulli parameter of routing policy $r_\phi(\mathbf{s}, x)$.
This cold-start IL phase aligns the coordinator’s latent space with the agents' performance differential, providing a stable policy initialization that significantly accelerates the subsequent RL phase.

\begin{figure*}[t]
    \centering
    \vspace{-1.0em}  
\includegraphics[width=\textwidth,height=8.5cm,keepaspectratio]{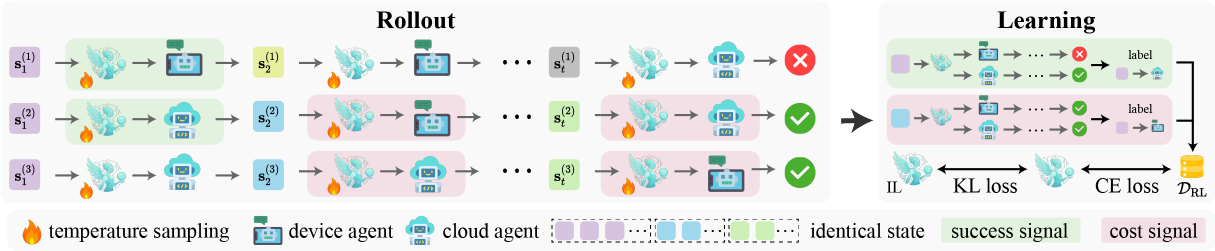}
\vspace{-1em}
\caption{RL workflow of \name. We group identical states in multiple rollouts, and derive cost-aware preference labels that favor higher return and fewer future cloud calls to update \name.}
\label{fig:hera_training_overview}
    \vspace{-1.5em}  
\end{figure*}

\subsection{Cost-Aware Reinforcement Learning with State Grouping}
\label{sec:hera_rl}
The IL phase provides a strong initialization but relies on a proxy objective and does not directly optimize Eq.~\ref{eq:opt_obj}. We then refine \name with reinforcement learning (RL)\footnote{We use trajectory return and cost as reinforcement signals, but optimize the router through induced step-level classification targets rather than direct policy gradients.}, jointly maximizing end-to-end task success while minimizing step-level cloud invocations.
We deploy device agent $\pi_{\theta_\mathrm{d}}$ and cloud agent $\pi_{\theta_\mathrm{c}}$, together with router $r_\phi$ trained in the preceding IL stage, into the environment and collect $N$ trajectories for each task $x$ from the same initial state.
To induce sufficient exploration, we sample routing decisions from a temperature-scaled Bernoulli distribution parameterized by the router's binary head.
Concretely, at step $t$ the router $r_\phi$ outputs a logit $\ell_\phi(\mathbf{s}_t,x)$, and we form:
\begin{equation}
p_t^{(\gamma)} \;=\; \sigma\!\left(\ell_\phi(\mathbf{s}_t,x)/\gamma\right),
\qquad
r_{\phi}(\mathbf{s}_t,x) \sim \mathrm{Bernoulli}\!\left(p_t^{(\gamma)}\right),
\label{eq:temperature}
\end{equation}
where temperature $\gamma>0$ (larger $\gamma$ increases stochasticity).
This yields $N$ trajectories $\{\tau_i\}_{i=1}^N$, where each is defined as
\(\tau_i = \{(\mathbf{s}_1^{(i)}, \hat{\mathbf{a}}_1^{(i)} ,r_1^{(i)}), (\mathbf{s}_2^{(i)}, \hat{\mathbf{a}}_2^{(i)} ,r_2^{(i)}), \dots, (\mathbf{s}_T^{(i)}, \hat{\mathbf{a}}_T^{(i)} ,r_T^{(i)})\}\). All trajectories share the same initial state, i.e., $\mathbf{s}_1^{(1)}=\mathbf{s}_1^{(2)}=,\cdots,=\mathbf{s}_1^{(N)}$.

We then group identical states to derive step-level learning signals.
Let $\mathcal{G}(x)$ denote the set of distinct states encountered across the $N$ trajectories.
For each grouped state $\mathbf{s}\in\mathcal{G}(x)$, we collect all its occurrences
$
\mathcal{J}(\mathbf{s}) \coloneqq \{\, (i,t) \mid \mathbf{s}^{(i)}_t=\mathbf{s} \,\},
$
which enumerates every visit to $\mathbf{s}$.
For each $\mathbf{s}\in\mathcal{G}(x)$, we estimate two quantities conditioned on the routing decision:
(1) the expected trajectory return, and
(2) the expected future cloud reliance.
Specifically, we compute decision-conditioned averages by restricting $r_{\phi} ({\mathbf{s}}_{t}^{(i)},x)=d$:
\begin{align}
\widehat{R}(d;\mathbf{s})
&:= \frac{1}{\sum_{(i,t)\in\mathcal{J}(\mathbf{s})}\mathbb{I}[r_{\phi} ({\mathbf{s}}_{t}^{(i)},x)=d]}
\sum_{(i,t)\in\mathcal{J}(\mathbf{s})}\mathbb{I}[r_{\phi} ({\mathbf{s}}_{t}^{(i)},x)=d]\; R(\tau_i), \label{eq:group_return2}\\
\widehat{C}(d;\mathbf{s})
&:= \frac{1}{\sum_{(i,t)\in\mathcal{J}(\mathbf{s})}\mathbb{I}[r_{\phi} ({\mathbf{s}}_{t}^{(i)},x)=d]}
\sum_{(i,t)\in\mathcal{J}(\mathbf{s})}\mathbb{I}[r_{\phi} ({\mathbf{s}}_{t}^{(i)},x)=d]\; C_{i,t}, \label{eq:group_cloud2}
\end{align}
for $d\in\{0,1\}$, where $d=0$ routes to $\pi_{\theta_\mathrm{d}}$ and $d=1$ routes to $\pi_{\theta_\mathrm{c}}$.
Here, $\widehat{R}(d;\mathbf{s})$ estimates the expected trajectory return, and $\widehat{C}(d;\mathbf{s})$ estimates the expected remaining cloud calls (including the current step), where
$C_{i,t} := \sum_{t' = t}^{T} \mathbb{I}\!\left[r_{\phi} ({\mathbf{s}}_{t'}^{(i)},x) = 1\right]$.

We use above estimates to derive a cost-aware routing rule for step-level supervision.
For each $\mathbf{s}\in\mathcal{G}(x)$, we define a preference label $y(\mathbf{s})\in\{0,1\}$ by selecting the decision with higher estimated return, and breaking near-ties by lower expected cloud usage.
Formally, with a margin $\epsilon>0$:
\begin{equation}
y(\mathbf{s}) :=
\begin{cases}
1, & \widehat{R}(1;\mathbf{s}) - \widehat{R}(0;\mathbf{s}) > \epsilon,\\
0, & \widehat{R}(0;\mathbf{s}) - \widehat{R}(1;\mathbf{s}) > \epsilon,\\
\arg\min\limits_{d\in\{0,1\}} \widehat{C}(d;\mathbf{s}), & \text{otherwise}.
\end{cases}
\label{eq:rl_label_rule}
\end{equation}
Here, $\epsilon$ controls the performance--cost trade-off: $\epsilon\!\to\!0$ favors higher $\widehat{R}$, while $\epsilon\!\to\!+\infty$ favors lower $\widehat{C}$.
In each RL iteration, we roll out $N$ trajectories and construct training set $\mathcal{D}_{\mathrm{RL}}=\{(\mathbf{s},x,y(\mathbf{s})) \mid \mathbf{s}\in\mathcal{G}(x)\}$, and update $r_{\phi}$  by minimizing the BCE loss over $\mathcal{D}_{\mathrm{RL}}$:
\begin{equation}
\mathcal{L}_{\mathrm{RL}}(\phi)
=
-
\frac{1}{|\mathcal{D}_{\mathrm{RL}}|}
\sum_{(\mathbf{s},x,y)\in\mathcal{D}_{\mathrm{RL}}}
\left[
y \log p_{\phi}(\mathbf{s},x)
+
(1-y)\log\big(1-p_{\phi}(\mathbf{s},x)\big)
\right]
+
\beta \, \|\phi-\phi_{\mathrm{IL}}\|_2^2,
\label{eq:rl_loss}
\end{equation}
where $\phi_{\mathrm{IL}}$ is the frozen router parameters obtained at the end of the imitation learning stage, and $\beta$ controls the regularization strength. This regularization stabilizes RL refinement by preventing \name from drifting excessively far from the well-initialized IL stage.

In general, the IL stage provides \name with a strong device--cloud coordination prior, and RL refines it into a cost-aware routing policy that directly optimizes task success under limited cloud usage. Together, \name achieves effective success--cost co-optimization in device--cloud LLM agent systems.

\begin{table}[t]
\centering
\caption{Task success rate, cloud LLM API cost, and end-to-end trajectory latency on ALFWorld, WebShop, and AppWorld, averaged over three runs.}
\label{tab:results}
\vspace{1mm}
\scriptsize
\setlength{\tabcolsep}{2.6pt}
\renewcommand{\arraystretch}{0.95}
\begin{tabular}{l ccc | ccc | ccc}
\toprule
\multirow{2}{*}{\textbf{Method}} 
& \multicolumn{3}{c|}{\textbf{ALFWorld}} 
& \multicolumn{3}{c|}{\textbf{WebShop}} 
& \multicolumn{3}{c}{\textbf{AppWorld}} \\
& Succ. (\%) & Cost ($10^{\scriptstyle -3}\,\$$) & Time (s) 
& Succ. & Cost & Time 
& Succ. & Cost & Time \\
\midrule
Device-only & $48.6_{\pm 1.9}$ & $0.0_{\pm 0.0}$ & $123.0_{\pm 9.1}$ 
          & $3.8_{\pm 0.2}$ & $0.0_{\pm 0.0}$ & $61.8_{\pm 2.5}$ 
          & $0.6_{\pm 0.2}$ & $0.0_{\pm 0.0}$ & $144.2_{\pm 7.4}$ \\
Cloud-only & $90.9_{\pm 0.8}$ & $11.7_{\pm 0.5}$ & $174.1_{\pm 4.9}$ 
           & $35.6_{\pm 1.0}$ & $14.7_{\pm 0.5}$ & $181.5_{\pm 6.3}$ 
           & $22.6_{\pm 1.6}$ & $43.8_{\pm 1.2}$ & $332.5_{\pm 10.3}$ \\
Rand. ($0.3$) & $72.1_{\pm 5.0}$ & $6.1_{\pm 0.2}$ & $157.6_{\pm 7.7}$ 
                              & $21.2_{\pm 1.3}$ & $5.8_{\pm 0.4}$ & $85.9_{\pm 3.4}$ 
                              & $2.4_{\pm 0.4}$ & $14.7_{\pm 0.3}$ & $304.1_{\pm 8.1}$ \\
Rand. ($0.4$) & $77.9_{\pm 4.7}$ & $7.2_{\pm 0.6}$ & $148.1_{\pm 10.5}$ 
                              & $24.5_{\pm 0.8}$ & $6.9_{\pm 0.6}$ & $97.6_{\pm 7.3}$ 
                              & $7.1_{\pm 0.5}$ & $19.2_{\pm 1.4}$ & $363.4_{\pm 13.2}$ \\
Rand. ($0.5$) & $79.1_{\pm 1.5}$ & $8.9_{\pm 1.1}$ & $169.5_{\pm 13.2}$ 
                              & $27.6_{\pm 1.1}$ & $8.3_{\pm 0.7}$ & $108.1_{\pm 4.2}$ 
                              & $5.4_{\pm 0.7}$ & $22.5_{\pm 1.1}$ & $392.7_{\pm 15.7}$ \\
Eagle    & $80.3_{\pm 3.6}$ & $8.5_{\pm 0.4}$ & $178.5_{\pm 6.3}$ 
         & $26.6_{\pm 3.0}$ & $8.0_{\pm 0.9}$ & $116.6_{\pm 2.9}$ 
         & $9.8_{\pm 1.4}$ & $18.2_{\pm 2.5}$ & $317.1_{\pm 14.4}$ \\
Hybrid~LLM  & $79.3_{\pm 1.1}$ & $7.9_{\pm 1.3}$ & $130.2_{\pm 5.8}$     & $29.1_{\pm 3.1}$ & $7.7_{\pm 0.8}$ & $109.3_{\pm 5.1}$ 
         & $16.2_{\pm 0.7}$ & $36.9_{\pm 2.3}$ & $301.9_{\pm 12.5}$ \\
RouteLLM & $83.6_{\pm 2.8}$ & $9.0_{\pm 0.8}$ & $156.7_{\pm 3.3}$
         & $31.1_{\pm 2.6}$ & $7.2_{\pm 0.3}$ & $155.1_{\pm 13.9}$ 
         & $17.6_{\pm 1.9}$ & $31.3_{\pm 4.2}$ & $266.3_{\pm 7.2}$ \\
FrugalGPT & $83.7_{\pm 1.4}$ & $8.7_{\pm 2.1}$ & $203.7_{\pm 4.7}$ 
         & $31.8_{\pm 1.3}$ & $12.1_{\pm 1.1}$ & $125.8_{\pm 8.8}$
         & $18.5_{\pm 2.7}$ & $37.5_{\pm 1.9}$ & $359.5_{\pm 19.3}$ \\
\cellcolor{blue!10}\name& \cellcolor{blue!10}$86.1_{\pm 3.2}$ & \cellcolor{blue!10}$4.0_{\pm 0.9}$ & \cellcolor{blue!10}$105.5_{\pm 5.3}$ 
         & \cellcolor{blue!10}$34.1_{\pm 2.2}$  & \cellcolor{blue!10}$8.1_{\pm 0.6}$ & \cellcolor{blue!10}$112.2_{\pm 4.4}$ 
         & \cellcolor{blue!10}$19.7_{\pm 1.3}$ & \cellcolor{blue!10}$31.3_{\pm 2.6}$ & \cellcolor{blue!10}$286.0_{\pm 9.8}$ \\
\bottomrule
\end{tabular}
\label{table:overall_text}
\vspace{-1em}  
\end{table}

\section{Experiment}
\label{sec:exp}

\vspace{-.5em}
We conduct comprehensive evaluations of \name across diverse agentic tasks to show:
(1) the strong device--cloud coordination capability of \name for an optimal performance--cost trade-off;
(2) the contribution of each training stage through ablation studies;
(3) microbenchmarks, including cloud offloading rate, compute overhead, and analyses of routing patterns.

\vspace{-.5em}
\subsection{Experimental Settings}
\label{subsec:exp_set}

\vspace{-.5em}
\paragraph{Benchmarks.}
We evaluate \name on three challenging long-horizon agent benchmarks:
ALFWorld~\citep{ALFWorld20}, WebShop~\citep{yao2022webshop}, and AppWorld~\citep{appworld}.
ALFWorld is an embodied benchmark with 3,827 tasks across six household activity categories, designed to evaluate multi-step decision-making and interactive control for LLM agents.
WebShop is a web-based benchmark for LLM agents on realistic shopping tasks, featuring over 1.1M products and 12k user instructions in a large, diverse action space.
AppWorld is a high-fidelity multi-app environment with 9 daily apps and 457 APIs, offering 750 interactive coding-centric agent tasks with robust programmatic evaluation via state- and execution-based unit tests.


\vspace{-.5em}
\paragraph{Baselines.}
We compare \name with a range of competitive baselines: 
(1) device-only and cloud-only LLM agents;
(2) stochastic step-level routing that randomly selects between device and cloud agents with predefined probabilities;
(3) state-of-the-art device--cloud LLM routing methods, including Eagle~\citep{Eagle2024}, FrugalGPT~\citep{chen2024frugalgpt}, Hybrid~LLM~\citep{ding2024hybrid}, and RouteLLM~\citep{ongroutellm}.
Eagle retrieves similar historical queries and combines global and local ELO scores~\citep{elo1967proposed} to route each request to the best LLM.
FrugalGPT first routes queries to a device-side model and escalates to a stronger cloud LLM only when a learned scorer deems the response unreliable.
Hybrid~LLM trains a lightweight router to predict query difficulty and dynamically assign each request to device or cloud model.
RouteLLM trains a win-prediction router on preference data and routes to the cloud model when its win probability exceeds a threshold.

\vspace{-.5em}
\paragraph{Implementation details.}
We use the Qwen2.5-7B-Instruct~\citep{yang2024qwen25} as the on-device model and the closed-source Qwen-Max API~\citep{qwen2.5-max2025} as the strong cloud oracle, respectively (other model pairs in Appendix~\ref{append:model_pair}).
\name is initialized from the Qwen2.5-0.5B backbone~\citep{yang2024qwen25} and replaces the LM head with a binary classification head, resulting in a 494M-parameter model.
Training is conducted on 2$\times$H100 GPUs, freezing the backbone and updating only the last two attention layers and the classification head (30M parameters).
We use AdamW~\citep{loshchilov2018decoupled} optimizer,
with learning rates of $4e-5$ (IL) and 1e$-$5 (RL), batch sizes of 64 (IL) and 256 (RL).
We set $N=8$, $\delta=0.5$, $\gamma=1.3$, $\epsilon=0.05$, and $\beta=0.1$.
After training, we use the MNN inference engine~\citep{alibaba2020mnn} to deploy both \name and device-side LLM on an NVIDIA GeForce RTX 4090 GPU, evaluating end-to-end performance and cost in a realistic device--cloud collaboration setting.
More training and evaluation details are provided in Appendix~\ref{append:exp_set}.


\vspace{-.5em}
\subsection{Overall Results}
\label{subsec:main_results}
Table~\ref{table:overall_text} reports the average task success rate and cost of \name and other baselines on ALFWorld, WebShop, and AppWorld benchmark, where cost includes closed-source cloud API fees and end-to-end latency.
The device-side agent yields the smallest latency but the lowest success rates.
By contrast, the cloud-only agent boosts success by 42.3\%/31.8\%/22.0\% on ALFWorld/WebShop/AppWorld, respectively, at a per-task cost of 11.7/14.7/43.8 $\times 10^{-3}$ dollars and a latency increase of 51.1/119.7/188.3 seconds.
A random router dispatches each step to the device or cloud with a fixed probability (cloud-routing probability in parentheses), yielding a naive success--cost trade-off; however, without step-difficulty or model-preference awareness, it is inefficient—raising the cloud-routing probability from 0.3 to 0.5 improves success by only 5.5\% on average while significantly increasing cloud calls, per-task cost, and latency.
Notably, reduced cloud-based long-horizon planning lengthens trajectories and introduces redundant steps.
The semantic-similarity-based router Eagle yields modest gains over random routing at comparable cloud usage, but still lags behind training-based methods in complex agentic environments.
Training-based baselines achieve similar results; FrugalGPT performs d better by conditioning routing on device outputs, but incurs extra latency by waiting for device generation.
By contrast, \name consistently outperforms all baselines, achieving higher task success rates under substantially lower cost.
It achieves 92.5\% of the cloud-only success rate with only 53.6\% of the cloud cost and 69.3\% of the end-to-end latency on average.
This gain is mainly attributed to the RL stage, whose end-to-end trajectory-level success--cost optimization enables \name to learn not only state-wise routing decisions but also their interdependence across consecutive steps, ultimately yielding more precise and cost-effective device--cloud coordination.

\begin{table}[t]
\centering
\caption{Ablation results for two-stage training of \name.}
\vspace{1mm}
\scriptsize
\renewcommand{\arraystretch}{0.95}
\begin{tabular}{l cc | cc | cc}
\toprule
\multirow{2}{*}{\textbf{Method}} 
& \multicolumn{2}{c|}{\textbf{ALFWorld}} 
& \multicolumn{2}{c|}{\textbf{WebShop}} 
& \multicolumn{2}{c}{\textbf{AppWorld}} \\
& Succ. (\%) & $\pi_{\theta_\mathrm{c}}$ calls
& Succ. & $\pi_{\theta_\mathrm{c}}$ calls 
& Succ. & $\pi_{\theta_\mathrm{c}}$ calls \\
\midrule
Rand. ($0.5$) & $79.1$
         & $14.5$  & $27.6$
         & $7.8$ & $5.4$ & $14.6$ \\
\name w/o RL & $79.9$ (\textcolor{green!50!black}{$\uparrow0.8$})  
         & $10.6$ (\textcolor{green!50!black}{$\downarrow3.9$})  & $31.1$ (\textcolor{green!50!black}{$\uparrow3.5$})  
         & $8.2$ (\textcolor{red!70!black}{$\uparrow0.4$})  & $9.2$ (\textcolor{green!50!black}{$\uparrow3.8$})  
         & $15.1$ (\textcolor{red!70!black}{$\uparrow0.5$}) \\
\name w/o IL  & $83.3$ (\textcolor{green!50!black}{$\uparrow4.2$}) 
         & $9.3$ (\textcolor{green!50!black} {$\downarrow5.2$}) & $29.9$ (\textcolor{green!50!black}{$\uparrow2.3$}) 
         & $7.2$ (\textcolor{green!50!black} {$\downarrow0.6$}) & $14.3$ (\textcolor{green!50!black}{$\uparrow8.9$}) & $17.3$ (\textcolor{red!70!black}{$\uparrow2.7$}) \\
\cellcolor{blue!10}\name 
         & \cellcolor{blue!10}$86.1$ (\textcolor{green!50!black}{$\uparrow7.0$})
         & \cellcolor{blue!10}$5.3$ (\textcolor{green!50!black}{$\downarrow9.2$}) & \cellcolor{blue!10}$34.1$ (\textcolor{green!50!black}{$\uparrow6.5$})
         & \cellcolor{blue!10}$5.1$ (\textcolor{green!50!black}{$\downarrow2.7$}) & \cellcolor{blue!10}$19.7$ (\textcolor{green!50!black}{$\uparrow14.3$}) & \cellcolor{blue!10}$15.7$ (\textcolor{red!70!black}{$\uparrow0.9$}) \\
\bottomrule
\end{tabular}
\label{table:abaltion}
\vspace{-2em}
\end{table}

\begin{wrapfigure}{rB}{0.715\textwidth}
\vspace{-1em}
    \centering
    \begin{subfigure}[t]{0.243\textwidth}
        \centering
        \includegraphics[width=\textwidth, valign=t]{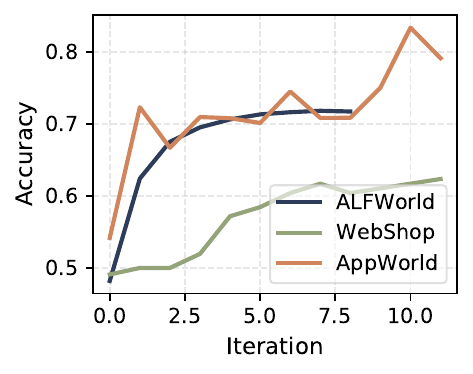}
        \vspace{-.7em}
        \caption{IL stage.}
        \label{subfig:il_progess}
    \end{subfigure}
    \begin{subfigure}[t]{0.222\textwidth}
        \centering
        \includegraphics[width=\textwidth, valign=t]{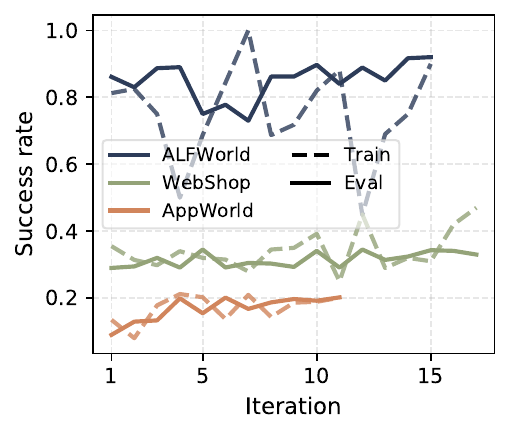}
        \vspace{-0.55em}
        \caption{RL stage (success).}
        \label{subfig:rl_success}
    \end{subfigure}
    \begin{subfigure}[t]{0.222\textwidth}
        \centering
        \includegraphics[width=\textwidth, valign=t]{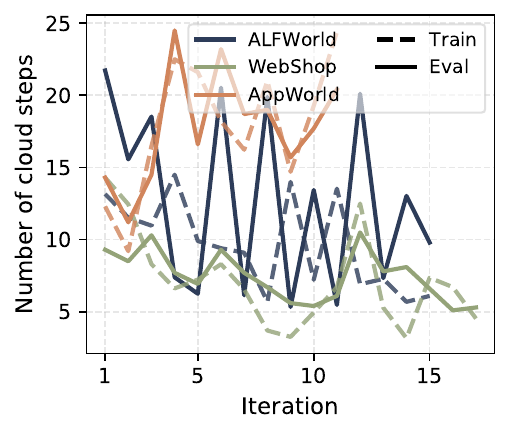}
        \vspace{-0.55em}
        \caption{RL stage (cost).}
        \label{subfig:rl_cost}
    \end{subfigure}
    \caption{Training dynamics of \name. IL: device--cloud consistency classification accuracy on a held-out validation set. RL: success rate and average cloud invocations on the benchmarks.}
    \label{fig:2_stage_curve}
    \vspace{-1em}
\end{wrapfigure}

\vspace{-.5em}
\subsection{Ablation Study}
\label{subsec:ablation}

Table~\ref{table:abaltion} reports benchmark results after ablating each training stage, including average per-task success rate and cloud-agent call count.
After the IL stage (w/o RL), \name already outperforms the random policy, improving success by 0.8\%/3.5\%/3.8\% on ALFWorld/WebShop/AppWorld while reducing cloud calls by 3.9/-0.4/-0.5 per task. This indicates that replaying cloud trajectories on the device model and using output-consistency supervision yields strong signals for detecting states that require cloud assistance, enabling effective coordination.
With RL refinement, \name further improves success by 6.6\% on average, while reducing cloud calls by 2.6 per task.
This highlights the effectiveness of our cost-aware RL design: in a complex state space, \name learns to identify the critical states where cloud assistance is most needed, achieving higher success with a minimal cloud-invocation budget.
Moreover, RL alone (w/o IL) performs worse than \name, since it lacks a good cold start, confirming that \name benefits from the two-stage IL$\rightarrow$RL training pipeline.


Figure~\ref{fig:2_stage_curve} further illustrates training dynamics across different stages of \name.
Figure~\ref{subfig:il_progess} shows the cold-start IL learning curve on a held-out validation split of $X_{\text{diff}}$.
Training $r_\phi$ on $\mathcal{D}_{\mathrm{IL}}$ improves routing accuracy to approximately 60\%--80\%, showing that device--cloud action consistency provides a strong and learnable coordination prior.
During RL refinement, \name leverages diverse trajectory rollouts to jointly optimize success and cost: trajectory success steadily improves (Figure~\ref{subfig:rl_success}), while cloud-invoked steps consistently decrease (Figure~\ref{subfig:rl_cost}), except in AppWorld where a slight increase in cloud calls yields substantial success gains.
As a results, \name achieve the highest success rate with the lowest cloud usage, demonstrating the effectiveness of carefully designed learning signals. Additional hyperparameter studies and ablations are provided in Appendix~\ref{appdendix:hyper}.

\vspace{-.5em}
\subsection{Micro-benchmarking}
This section presents micro-benchmarks of \name from multiple perspectives, ncluding the number of cloud-invoked steps, computational overhead, and routing patterns.

\begin{figure}[t]
\centering
\begin{minipage}[t]{0.73\textwidth}
  \centering
  \begin{subfigure}[t]{0.32\textwidth}
    \includegraphics[width=\textwidth]{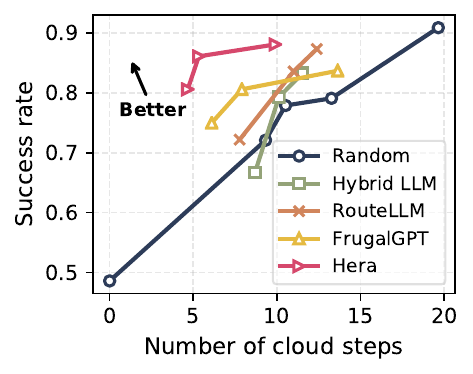}
    \caption{ALFWorld.}
  \end{subfigure}
  \begin{subfigure}[t]{0.32\textwidth}
    \includegraphics[width=\textwidth]{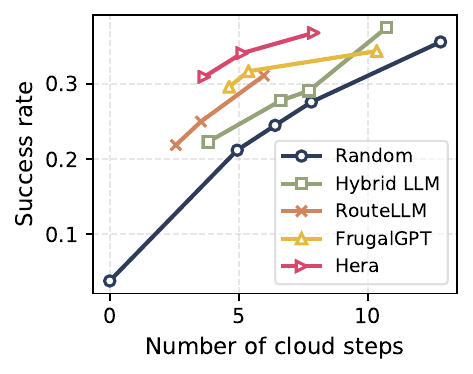}
    \caption{WebShop.}
  \end{subfigure}
  \begin{subfigure}[t]{0.32\textwidth}
    \includegraphics[width=\textwidth]{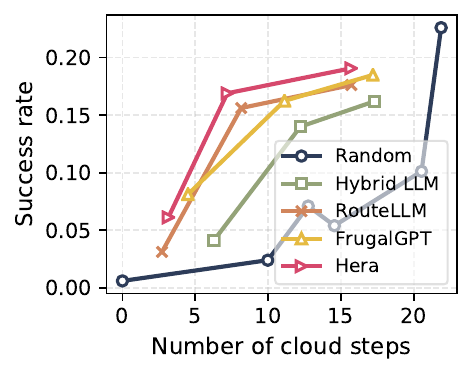}
    \caption{AppWorld.}
  \end{subfigure}
  \caption{Task success rate of \name and baselines as a function of cloud invocation count on ALFWorld, WebShop, and AppWorld.}
  \label{fig:cloud_assist}
\end{minipage}\hfill
\begin{minipage}[t]{0.25\textwidth}
  \centering
  \includegraphics[width=\textwidth]{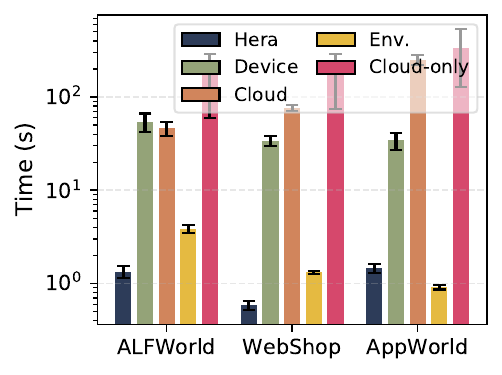}
  \vspace{-1.7em}
  \caption{End-to-end latency breakdown for device--cloud collaborative LLM agent systems.}
  \label{fig:latency_break}
\end{minipage}
\vspace{-1.5em}
\end{figure}

\paragraph{Number of cloud steps.}
Figure~\ref{fig:cloud_assist} shows the task success rates of \name and baselines with varying number of cloud-invoked steps.
Under random routing, success grows approximately linearly with the number of steps executed by $\pi_{\theta_\mathrm{c}}$ in ALFWorld and WebShop; interestingly, AppWorld exhibits a noticeable improvement after about 21 cloud calls per trajectory, reflecting its stronger reliance on the cloud model’s coding capability.
Training-based baselines (HybridLLM, RouteLLM, FrugalGPT) outperform random routing but still trail \name, as they rely on static estimates—akin to our IL stage—whereas \name’s cost-aware RL directly optimizes the success--cost trade-off from a holistic trajectory perspective. In comparison, \name achieves superior success rates at equivalent cloud costs and establish a strong Pareto frontier in device--cloud LLM agent systems.

\vspace{-.5em}
\paragraph{Computational overhead.}
Figure~\ref{fig:latency_break} breaks down per-task latency when \name coordinates the device agent on an NVIDIA RTX 4090 GPU with the cloud-model API, revealing that the end-to-end execution time is predominantly dominated by device and cloud LLM inference.
Notably, deploying \name on device incurs only marginal overhead, adding merely 61ms per step.
As a result, \name attains 95\%/96\%/87\% of cloud-only success on ALFWorld/WebShop/AppWorld, respectively, while incurring only 60\%/62\%/86\% of cloud execution latency via efficient device--cloud coordination.
Thus, \name proves to be exceptionally lightweight; the substantial gains in task success and the overall latency reduction achieved by accurately routing steps to the device outweigh the trivial on-device computational overhead, validating its suitability for real-world, latency-sensitive deployments.

\begin{wrapfigure}{r}{0.52\textwidth}
\vspace{-1em}
    \centering
    \begin{subfigure}[t]{0.235\textwidth}
        \centering
        \includegraphics[width=\textwidth, valign=t]{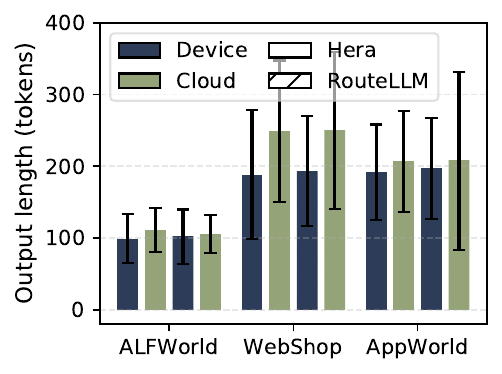}
        \caption{Reasoning length.}
        \label{subfig:333}
    \end{subfigure}
    \begin{subfigure}[t]{0.235\textwidth}
        \centering
        \includegraphics[width=\textwidth, valign=t]{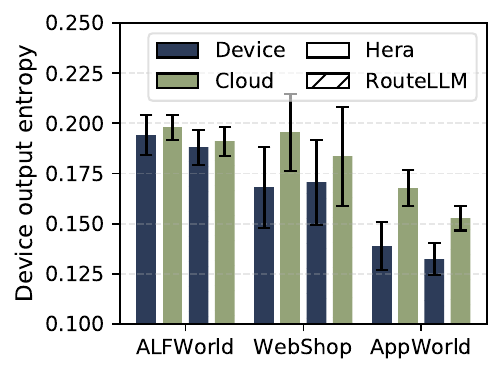}
        \caption{Predictive entropy.}
        \label{subfig:ent_bar_exp}
    \end{subfigure}
    \caption{Device-side agent output characteristics for device- and cloud-executed steps.}
    \label{fig:moti_exp_stepwise}
\vspace{-1em}
\end{wrapfigure}

\vspace{-.5em}
\paragraph{Routing patterns.}
Following \S~\ref{sec:motivation}, we compare \name with RouteLLM in terms of reasoning length and predictive entropy (replayed on $\pi_{\theta_\mathrm{d}}$) for states with distinct routing decisions in successful trajectories, demonstrating its implicit capture of task difficulty.
Compared to RouteLLM, \name induces a more pronounced separation in both reasoning length and entropy between device- and cloud-executed steps when evaluated on the device agent.
This sharper divergence, together with the success--cost results, indicates that \name more effectively identifies high-uncertainty, cognitively demanding steps, enabling more discriminative and difficulty-aware step-level routing.

\section{Conclusions}

This paper proposes \name, a step-level device--cloud LLM agent coordinator for long-horizon tasks. Specifically, \name accurately identifies whether cloud assistance is needed at each interaction state, thus enabling effective task execution with minimal cloud invocations. \name adopts a two-stage training paradigm. First, we perform cold-start imitation learning by replaying the device agent on cloud trajectories and using step-level action consistency as supervision, providing \name with an initial device--cloud routing capability.
Next, we refine \name via cost-aware reinforcement learning: across multiple rollouts, we group identical states and generate step-level preferences that favor higher return and, when comparable, fewer future cloud calls, thereby explicitly co-optimizing success and cost.
Extensive experiments on ALFWorld, WebShop, and AppWorld demonstrate that \name consistently outperforms state-of-the-art baselines while invoking the cloud agent in only 46.3\% of steps and achieving 92.5\% of cloud-only success.
We present the broader impact of \name in Appendix~\ref{appendix:impact}, and discuss the limitations of \name and directions for future research in Appendix~\ref{appdendix:limit_future}.

\begin{ack}
This work was supported by Key Research and Development Program of Zhejiang Province (No. 2025C01026)
\end{ack}


{\small
\bibliographystyle{plainnat}
\bibliography{refs}
}


\newpage
\appendix


\section{Broader Impacts}
\label{appendix:impact}

This work introduces \name as the first step-level device--cloud coordinator for LLM agents, marking a pivotal shift from coarse, task-level routing to fine-grained, dynamic collaboration. By demonstrating that only a small fraction of critical steps require cloud intervention, \name significantly lowers the computational and financial barriers to deploying sophisticated agents on user devices. This efficiency democratizes access to advanced agentic capabilities, enabling real-time, latency-sensitive applications—such as autonomous robotic control and interactive assistants—to operate reliably on device hardware without prohibitive costs.

Amid the explosive growth of LLM agents, artificial intelligence is undergoing a fundamental paradigm shift from static, single-turn queries to long-horizon autonomous interaction, where agents must sustain coherent reasoning over extended, dynamic trajectories. We believe \name is essential for realizing scalable, sustainable, and ubiquitous autonomous agents in real-world applications by establishing a strong performance--cost Pareto frontier for this complex mode of operation.

\section{Pseudocode}
\label{append:code}

Algorithm~\ref{alg:hera} summarizes the 2-stage training procedure of \name.

\begin{algorithm}[h]
\caption{\name:Two-Stage Training for Step-Level Device--Cloud LLM Agent Coordination}
\label{alg:hera}
\begin{algorithmic}[1]
\STATE \textbf{Input:} Device policy $\pi_{\theta_d}$, cloud policy $\pi_{\theta_c}$, router $r_\phi$; margins $\delta,\epsilon$; temperature $\gamma$; rollouts per task $N$; regularization weight $\beta$.
\STATE \textbf{Output:} Trained router parameters $\phi$.

\STATE {\bf Stage I: Cold-start imitation learning}
\STATE Initialize $\mathcal{D}_{\mathrm{IL}} \leftarrow \emptyset$.
\FORALL{tasks $x$}
    \STATE Roll out $\pi_{\theta_d}$ to get $\tau_d(x)$; roll out $\pi_{\theta_c}$ to get $\tau_c(x)$.
    \IF{$R(\tau_c(x)) - R(\tau_d(x)) > \delta$}
        \FOR{$t = 1,2,\ldots,T$}
            \STATE Let $(s_t^c,a_t^c)$ be the $t$-th state--action pair in $\tau_c(x)$.
            \STATE Sample $a_t^d \sim \pi_{\theta_d}(\cdot \mid s_t^c,x)$.
            \STATE $y_t \leftarrow \mathbb{I}[a_t^d = a_t^c]$.
            \STATE $\mathcal{D}_{\mathrm{IL}} \leftarrow \mathcal{D}_{\mathrm{IL}} \cup \{(s_t^c,x,y_t)\}$.
        \ENDFOR
    \ENDIF
\ENDFOR
\STATE Update $\phi$ by minimizing the IL objective $\mathcal{L}_{\mathrm{IL}}(\phi)$ (Eq.~\ref{eq:il}).
\STATE Save $\phi_{\mathrm{IL}} \leftarrow \phi$.

\STATE {\bf Stage II: Cost-aware reinforcement learning with state grouping}
\FOR{each RL iteration}
    \STATE Initialize $\mathcal{D}_{\mathrm{RL}} \leftarrow \emptyset$.
    \FORALL{tasks $x$}
        \STATE Collect $N$ trajectories $\{\tau_i\}_{i=1}^N$ using temperature $\gamma$ (Eq.~\ref{eq:temperature}).
        \STATE Group identical states across $\{\tau_i\}$ to obtain $\mathcal{G}(x)$ and $J(s)$.
        \FORALL{$s \in \mathcal{G}(x)$}
            \STATE Estimate $\widehat{R}(d;s)$ and $\widehat{C}(d;s)$ for $d\in\{0,1\}$ (Eqs.~\ref{eq:group_return2} and~\ref{eq:group_cloud2}).
            \STATE Derive preference label $y(s)$ (Eq.~\ref{eq:rl_label_rule}).
            \STATE $\mathcal{D}_{\mathrm{RL}} \leftarrow \mathcal{D}_{\mathrm{RL}} \cup \{(s,x,y(s))\}$.
        \ENDFOR
    \ENDFOR
    \STATE Update $\phi$ by minimizing $\mathcal{L}_{\mathrm{RL}}(\phi)$ with regularization to $\phi_{\mathrm{IL}}$ (Eq.~\ref{eq:rl_loss}).
\ENDFOR

\RETURN $\phi$.
\end{algorithmic}
\end{algorithm}

\section{Experimental Details}
\label{appendix:exp}

\subsection{Prompts}
The prompts used by LLM agents in ALFWorld, WebShop, and AppWorld are shown below.

\begin{promptbox}[title=Prompt Template for ALFWorld]
\begin{lstlisting}[style=mytt]
You are an expert agent operating in the ALFRED Embodied Environment. Your task is to: {task_description}
Prior to this step, you have already taken {step_count} step(s). Below are the most recent {history_length} observations and the corresponding actions you took: {action_history}
You are now at step {current_step} and your current observation is: {current_observation}
Your admissible actions of the current situation are: [{admissible_actions}].

Now it's your turn to take an action.
You should first reason step-by-step about the current situation. This reasoning process MUST be enclosed within <think> </think> tags. 
Once you've finished your reasoning, you should choose an admissible action for current step and present it within <action> </action> tags.
\end{lstlisting}
\end{promptbox}

\begin{promptbox}[title=Prompt Template for WebShop]
\begin{lstlisting}[style=mytt]
You are an expert autonomous agent operating in the WebShop e-commerce environment.
Your task is to: {task_description}.
Prior to this step, you have already taken {step_count} step(s). Below are the most recent
{history_length} observations and the corresponding actions you took: {action_history}
You are now at step {current_step} and your current observation is: {current_observation}.
Your admissible actions of the current situation are:
[
{available_actions}
].

Now it's your turn to take one action for the current step.
You should first reason step-by-step about the current situation, then think carefully which
admissible action best advances the shopping goal. This reasoning process MUST be enclosed
within <think> </think> tags.
Once you've finished your reasoning, you should choose an admissible action for current step
and present it within <action> </action> tags.
\end{lstlisting}
\end{promptbox}

\begin{promptbox}[title=Prompt Template for AppWorld]
\begin{lstlisting}[style=mytt]
I am your supervisor and you are a super intelligent AI Assistant whose job is to achieve my day-to-day tasks completely autonomously.

To do this, you will need to interact with app/s (e.g., spotify, venmo, etc) using their associated APIs on my behalf. For this you will undertake a *multi-step conversation* using a python REPL environment. That is, you will write the python code and the environment will execute it and show you the result, based on which, you will write python code for the next step and so on, until you've achieved the goal. This environment will let you interact with app/s using their associated APIs on my behalf.

Here are three key APIs that you need to know to get more information

# To get a list of apps that are available to you.
print(apis.api_docs.show_app_descriptions())

# To get the list of apis under any app listed above, e.g. supervisor
print(apis.api_docs.show_api_descriptions(app_name='supervisor'))

# To get the specification of a particular api, e.g. supervisor app's show_account_passwords
print(apis.api_docs.show_api_doc(app_name='supervisor', api_name='show_account_passwords'))

Each code execution will produce an output that you can use in subsequent calls. Using these APIs, you can now generate code, that the environment will execute, to solve the task.

Using these APIs, now begin writing code cells step-by-step to solve the actual task:

My name is: {supervisor_first_name} {supervisor_last_name}. My personal email is {supervisor_email} and phone number is {supervisor_phone_number}.

Your task is: {task_description}

Now it's your turn to generate code to solve the task.
You should first reason step-by-step about which APIs to call, what arguments to use, and how to build your code block to complete the task. This reasoning process MUST be enclosed within <think> </think> tags.
Once you've finished your reasoning, you present the solution code body within <code> </code> tags.
\end{lstlisting}
\end{promptbox}

\subsection{Details of Experimental Settings}
\label{append:exp_set}

For ALFWorld benchamrks ,we use a maximum prompt length of 2048 tokens and a maximum response length of 512 tokens.
Each episode is capped at 50 environment interaction steps, counting both device- and cloud-executed actions.
In WebShop, the prompt length is capped at 4096 tokens and the response length at 1024 tokens. Each episode terminates after 20 interaction steps.
In AppWorld, we set the maximum prompt length to 4096 tokens, the maximum response length to 1024 tokens, and the maximum number of steps to 50.
At each execution step, the agent is given access to a bounded history window. The number of visible historical steps is set to 50, 15, and 5 for the three datasets, respectively.
To evaluate request-level baselines in long-horizon agent settings, we apply them independently at each interaction step. Specifically, at step $t$, each baseline receives the same per-step context as \name—task instruction, current observation, and truncated history—and outputs a binary decision for routing the current step to either the device or the cloud model.

The definition of an identical state in the RL training stage depends on the benchmark-specific environment representation. In ALFWorld, two states are considered identical if they correspond to the same textual environment observation and admissible action set. In WebShop, two states are considered identical if they have the same webpage observation, including the current page content and available actions. In AppWorld, two states are considered identical if they produce the same execution observation returned by the Python environment under the same task context.

For device-side model deployment, we run the open-source MNN engine~\citep{alibaba2020mnn} on a single NVIDIA RTX 4090 GPU using BF16 precision. Cloud-side closed-source models are invoked exclusively through their official APIs. All training experiments are performed on 2 NVIDIA H100 GPUs. Unless otherwise specified, the imitation learning and reinforcement learning phases are run for 30 and 50 iterations, respectively. In practice, however, we observe that the best-performing checkpoint typically emerges much earlier, after only around 10 IL iterations and 15 RL iterations, highlighting the training efficiency of \name.


\subsection{Generalization across Device--Cloud Model Pairs}
\label{append:model_pair}
We further validate \name’s generalizability across diverse device--cloud model pairs on ALFWorld benchmark.
Beyond the default pairing of Qwen2.5-7B-Instruct (device) and Qwen-Max (cloud), we additionally use Llama 3.1 8B Instruct~\citep{llama31} as an alternative device model and GPT-5.4~\citep{gpt5-4} as a cloud model, and evaluate all device--cloud combinations formed from these models.
The results in Table~\ref{table:model_pair} demonstrate that \name consistently outperforms baselines across all model pairs, maintaining a superior performance--cost trade-off regardless of the underlying LLM architectures. While absolute success rates and costs vary depending on the capability gap between the specific device and cloud models, \name effectively adapts its routing policy to exploit the complementary strengths of each pair. This confirms that \name’s two-stage training paradigm is model-agnostic and robust, enabling flexible deployment in heterogeneous device--cloud collaborative environments.

\begin{table}[h]
\centering
\caption{Task success rate, cloud LLM API cost, and end-to-end trajectory latency on ALFWorld across different device--cloud model pairs, averaged over three runs.}
\vspace{1mm}
\scriptsize
\setlength{\tabcolsep}{2.6pt}
\renewcommand{\arraystretch}{0.95}
\begin{tabular}{l ccc | ccc | ccc}
\toprule
\multirow{2}{*}{\textbf{Method}} 
& \multicolumn{3}{c|}{\textbf{Qwen2.5 7B ($\pi_{\theta_\mathrm{d}}$) and GPT‑5.4 ($\pi_{\theta_\mathrm{c}}$)}} 
& \multicolumn{3}{c|}{\textbf{Llama 3.1 8B and Qwen-Max}} 
& \multicolumn{3}{c}{\textbf{Llama 3.1 8B and GPT-5.4}} \\
& Succ. (\%) & Cost ($10^{\scriptstyle -3}\,\$$) & Time (s) 
& Succ. & Cost & Time 
& Succ. & Cost & Time \\
\midrule
Device-only & $48.6_{\pm 1.9}$ & $0.0_{\pm 0.0}$ & $123.0_{\pm 9.1}$ 
          & $28.9_{\pm 1.2}$ & $0.0_{\pm 0.0}$ & $141.0_{\pm 7.3}$ 
          & $28.9_{\pm 1.2}$ & $0.0_{\pm 0.0}$ & $141.0_{\pm 7.3}$ \\
Cloud-only & $93.6_{\pm 0.6}$ & $68.8_{\pm 3.1}$ & $151.1_{\pm 13.2}$ 
           & $90.9_{\pm 0.8}$ & $11.7_{\pm 0.5}$ & $174.1_{\pm 14.9}$  
           & $93.6_{\pm 0.6}$ & $68.8_{\pm 3.1}$ & $151.1_{\pm 13.2}$ \\
Rand. ($0.3$) & $73.2_{\pm 4.4}$ & $35.0_{\pm 5.1}$ & $141.2_{\pm 8.6}$
                              & $47.2_{\pm 1.8}$ & $6.5_{\pm 0.4}$ & $149.0_{\pm 10.4}$ 
                              & $51.3_{\pm 2.9}$ & $31.0_{\pm 2.4}$ & $144.2_{\pm 10.1}$ \\
Rand. ($0.4$) & $78.1_{\pm 4.9}$ & $45.6_{\pm 2.7}$ & $149.7_{\pm 5.9}$
                              & $58.5_{\pm 0.7}$ & $7.1_{\pm 0.7}$ & $155.1_{\pm 7.4}$ 
                              & $63.3_{\pm 5.2}$ & $38.8_{\pm 2.6}$ & $133.3_{\pm 5.9}$ \\
Rand. ($0.5$) & $79.0_{\pm 2.2}$ & $47.8_{\pm 4.6}$ & $153.8_{\pm 8.8}$
                              & $62.3_{\pm 3.1}$ & $9.1_{\pm 0.9}$ & $164.6_{\pm 6.2}$ 
                              & $69.9_{\pm 4.8}$ & $44.3_{\pm 4.5}$ & $142.2_{\pm 15.3}$ \\
Eagle    & $81.2_{\pm 6.5}$ & $50.4_{\pm 6.3}$ & $161.2_{\pm 13.9}$
         & $75.7_{\pm 2.9}$ & $8.4_{\pm 1.1}$ & $175.3_{\pm 6.0}$ 
         & $73.2_{\pm 4.9}$ & $41.7_{\pm 3.8}$ & $162.3_{\pm 9.0}$ \\
Hybrid~LLM   & $76.2_{\pm 4.7}$ & $46.3_{\pm 2.9}$ & $129.2_{\pm 8.5}$ 
         & $76.8_{\pm 4.1}$ & $8.1_{\pm 0.8}$ & $127.7_{\pm 4.4}$ 
         & $76.7_{\pm 3.6}$ & $39.3_{\pm 7.7}$ & $129.4_{\pm 7.1}$ \\
RouteLLM  & $85.5_{\pm 3.1}$ & $52.3_{\pm 4.3}$ & $144.7_{\pm 12.1}$
         & $81.8_{\pm 3.5}$ & $8.9_{\pm 1.0}$ & $145.3_{\pm 7.1}$ 
         & $85.2_{\pm 2.5}$ & $37.2_{\pm 5.2}$ & $172.6_{\pm 16.8}$ \\
FrugalGPT & $88.6_{\pm 4.0}$ & $47.2_{\pm 2.9}$ & $195.5_{\pm 19.2}$
         & $82.4_{\pm 4.0}$ & $8.3_{\pm 0.7}$ & $210.3_{\pm 15.9}$ 
         & $87.3_{\pm 3.7}$ & $36.2_{\pm 1.7}$ & $199.3_{\pm 14.3}$ \\
\cellcolor{blue!10}\name 
         & \cellcolor{blue!10}$89.3_{\pm 2.6}$ & \cellcolor{blue!10}$23.5_{\pm 4.0}$ & \cellcolor{blue!10}$107.8_{\pm 7.8}$ 
         & \cellcolor{blue!10}$83.3_{\pm 5.4}$ & \cellcolor{blue!10}$5.3_{\pm 0.3}$ & \cellcolor{blue!10}$101.2_{\pm 7.6}$ 
         & \cellcolor{blue!10}$88.5_{\pm 4.2}$ & \cellcolor{blue!10}$26.8_{\pm 2.2}$ & \cellcolor{blue!10}$129.6_{\pm 9.5}$ \\
\bottomrule
\end{tabular}
\label{table:model_pair}
\vspace{-1em}  
\end{table}

\subsection{Effect of Hyperparameters}
\label{appdendix:hyper}
We investigate the sensitivity of \name to the  hyperparameter $\epsilon$ and $\beta$. $\epsilon$ governs the trade-off between task success and cloud cost in the RL stage: a small $\epsilon$ (e.g., 0) prioritizes success but incurs excessive cloud usage, while a large $\epsilon$ (e.g., 0.3) overly restricts offloading, leading to performance degradation. Similarly, $\beta$ stabilizes RL updates against the IL initialization; removing regularization ($\beta=0$) causes policy drift, whereas excessive regularization ($\beta$=0.3) hinders adaptation to RL objectives.
Empirical results in Table~\ref{table:hyper} indicate that the optimal configuration ($\epsilon=0.05$, $\beta=0.1$) effectively balances these dynamics, allowing \name to refine the IL prior into a robust, cost-efficient routing policy. This demonstrates that our design successfully leverages moderate regularization to prevent catastrophic forgetting while utilizing a fine-grained preference margin to identify critical states for cloud intervention, ensuring superior co-optimization of accuracy and efficiency.


\begin{table}[h]
\centering
\caption{Hyperparameter sensitivity of \name on ALFWorld.}
\vspace{1mm}
\scriptsize
\renewcommand{\arraystretch}{0.95}
\begin{tabular}{l cc | cc | cc}
\toprule
\multirow{2}{*}{\textbf{Method}} 
& \multicolumn{2}{c|}{\textbf{ALFWorld}} 
& \multicolumn{2}{c|}{\textbf{WebShop}} 
& \multicolumn{2}{c}{\textbf{AppWorld}} \\
& Succ. (\%) & $\pi_{\theta_\mathrm{c}}$ calls
& Succ. & $\pi_{\theta_\mathrm{c}}$ calls 
& Succ. & $\pi_{\theta_\mathrm{c}}$ calls \\
\midrule
\name ($\epsilon=0$, $\beta=0.1$) & $92.6$ (\textcolor{green!50!black}{$\uparrow6.5$})
         & $11.7$ (\textcolor{red!70!black}{$\uparrow6.6$})  & $38.1$ (\textcolor{green!50!black}{$\uparrow12.3$})
         & $15.2$ (\textcolor{red!70!black}{$\uparrow9.9$}) & $21.3$ (\textcolor{green!50!black}{$\uparrow1.6$})
         & $19.4$ (\textcolor{red!70!black}{$\uparrow3.7$}) \\
\name ($\epsilon=0.3$, $\beta=0.1$) & $65.7$ (\textcolor{red!70!black}{$\downarrow20.4$})
         & $2.4$ (\textcolor{green!50!black}{$\downarrow2.9$})   & $11.2$ (\textcolor{red!70!black}{$\downarrow22.9$})
         & $1.6$ (\textcolor{green!50!black}{$\downarrow3.5$}) & $15.1$ (\textcolor{red!70!black}{$\downarrow4.6$})
         & $10.5$ (\textcolor{green!50!black}{$\downarrow5.3$}) \\
\name ($\epsilon=0.05$, $\beta=0$) & $84.0$ (\textcolor{red!70!black}{$\downarrow2.1$})  
         & $7.7$ (\textcolor{red!70!black}{$\uparrow2.4$})  & $31.9$ (\textcolor{red!70!black}{$\downarrow2.2$})  
         & $4.9$ (\textcolor{green!50!black}{$\downarrow0.2$}) & $13.2$ (\textcolor{red!70!black}{$\downarrow6.5$})  
         & $12.1$ (\textcolor{red!70!black}{$\downarrow3.6$})   \\
\name ($\epsilon=0.05$, $\beta=0.3$)  &$79.9$ (\textcolor{red!70!black}{$\downarrow6.2$})  
         & $4.9$ (\textcolor{green!50!black}{$\downarrow0.4$})  & $29.8$ (\textcolor{red!70!black}{$\downarrow4.3$})  
         & $4.6$ (\textcolor{green!50!black}{$\downarrow0.5$}) & $17.7$ (\textcolor{red!70!black}{$\downarrow2.0$})  
         & $16.2$ (\textcolor{red!70!black}{$\uparrow0.5$})  \\
\cellcolor{blue!10}\name ($\epsilon=0.05$, $\beta=0.1$)
         & \cellcolor{blue!10}$86.1$ 
         & \cellcolor{blue!10}$5.3$  & \cellcolor{blue!10}$34.1$
         & \cellcolor{blue!10}$5.1$ & \cellcolor{blue!10}$19.7$ & \cellcolor{blue!10}$15.7$ \\
\bottomrule
\end{tabular}
\label{table:hyper}
\end{table}

\section{Case Study}
\label{appendix:case}
This section presents case studies to investigate how \name performs step-level device--cloud coordination in agent tasks.
Concretely, we analyze representative successful trajectories on three benchmarks—ALFWorld, WebShop, and AppWorld—each exhibiting distinct structural properties in terms of action space, reasoning patterns, and feedback signals. For each benchmark, we highlight critical decision points, compare device and cloud behaviors, and demonstrate how \name selectively invokes the cloud only when necessary to preserve task success while minimizing cost.

\subsection{ALFWorld}
The example shows how \name coordinates device- and cloud-side agents on ALFWorld. For routine exploration, the device closely matches the cloud, making local execution sufficient. For critical steps requiring state tracking and long-horizon consistency, the cloud becomes decisive: in Step 19, it avoids the device’s redundant exploration and advances the task. This shows that \name learns to invoke the cloud only for success-critical decisions.

\begin{envbox}[title=Step 1: environment]
\begin{lstlisting}[style=mytt]
-= Welcome to TextWorld, ALFRED! =-
You are in the middle of a room. Looking quickly around you, you see a cabinet 20, a cabinet 19, a cabinet 18, a cabinet 17, a cabinet 16, a cabinet 15, a cabinet 14, a cabinet 13, a cabinet 12, a cabinet 11, a cabinet 10, a cabinet 9, a cabinet 8, a cabinet 7, a cabinet 6, a cabinet 5, a cabinet 4, a cabinet 3, a cabinet 2, a cabinet 1, a coffeemachine 1, a countertop 2, a countertop 1, a diningtable 2, a diningtable 1, a drawer 6, a drawer 5, a drawer 4, a drawer 3, a drawer 2, a drawer 1, a fridge 1, a garbagecan 1, a microwave 1, a sinkbasin 1, a stoveburner 4, a stoveburner 3, a stoveburner 2, a stoveburner 1, and a toaster 1.
Your task is to: cool some winebottle and put it in diningtable.
\end{lstlisting}
\end{envbox}

\begin{herabox}[title=Step 1: Hera decision]
\begin{lstlisting}[style=mytt]
0 (device)
\end{lstlisting}
\end{herabox}

\begin{devicebox}[title=Step 1: device agent output]
\begin{lstlisting}[style=mytt]
<think>
To complete the task of cooling some wine bottle and putting it on the dining table, I need to first locate a wine bottle. Given the options, the most likely place to find a wine bottle would be in one of the cabinets, as they are often used for storing food and beverages. I will start by checking the cabinets, beginning with cabinet 1.
</think>
<action>
go to cabinet 1
</action>
\end{lstlisting}
\end{devicebox}

\begin{cloudbox}[title=Step 1: cloud agent output ((if routed)]
\begin{lstlisting}[style=mytt]
<think>
The task is to cool a winebottle and then place it on the diningtable. The first step would be to find a winebottle. Winebottles are often stored in cabinets, so I should check the cabinets. After finding the winebottle, I will need to cool it, which can be done using the fridge. Finally, after cooling, I will take the winebottle to one of the diningtables.
Since there are multiple cabinets, I don't know which one contains the winebottle. I will start by checking cabinet 1 and proceed sequentially if necessary.
</think>
<action>
go to cabinet 1
</action>
\end{lstlisting}
\end{cloudbox}

\begin{envbox}[title=Step 19: environment]
\begin{lstlisting}[style=mytt]
You move the glassbottle 1 to the diningtable 1.
\end{lstlisting}
\end{envbox}

\begin{herabox}[title=Step 19: Hera decision]
\begin{lstlisting}[style=mytt]
1 (cloud)
\end{lstlisting}
\end{herabox}

\begin{cloudbox}[title=Step 19: cloud agent output]
\begin{lstlisting}[style=mytt]
<think>
The task is to cool a winebottle and then place it on the diningtable. From the previous observations, we found that there is already a winebottle 1 on the diningtable 1. The next step would be to take this winebottle 1 from the diningtable 1 and move it to the fridge 1 to cool it down. Once it's cooled, we can then bring it back to the diningtable 1. Since I am currently at the diningtable 1 and have moved the glassbottle 1 there, my immediate action should be to take the winebottle 1.
</think>
<action>
take winebottle 1 from diningtable 1
</action>
\end{lstlisting}
\end{cloudbox}

\begin{devicebox}[title=Step 19: device agent output (if routed)]
\begin{lstlisting}[style=mytt]
<think>
1. I have moved the glassbottle 1 to the diningtable 1.
2. However, there was no winebottle on the diningtable 1 during the last observation.
3. The task requires me to cool a winebottle and put it on the diningtable.
4. Since the winebottle is not on the diningtable, I need to find it elsewhere in the environment.
5. The winebottle could be in one of the cabinets or on another countertop.
6. I will check the cabinets and countertops systematically to find the winebottle.
</think>
<action>
go to cabinet 15
</action>
\end{lstlisting}
\end{devicebox}

\begin{envbox}[title=Step 24: environment]
\begin{lstlisting}[style=mytt]
You arrive at diningtable 1. On the diningtable 1, you see a bowl 2, a bread 1, a butterknife 1, a dishsponge 1, a glassbottle 1, a kettle 1, a spatula 2, a spatula 1, and a tomato 1.
\end{lstlisting}
\end{envbox}

\begin{herabox}[title=Step 24: Hera decision]
\begin{lstlisting}[style=mytt]
0 (device)
\end{lstlisting}
\end{herabox}

\begin{devicebox}[title=Step 24: device agent output]
\begin{lstlisting}[style=mytt]
<think>
I need to cool the winebottle and then put it on the diningtable. From the previous steps, I have already cooled the winebottle 1 using the fridge 1. Now, I need to place it back on the diningtable 1. The action "move winebottle 1 to diningtable 1" is available and appropriate for this task.
</think>
<action>
move winebottle 1 to diningtable 1
</action>
\end{lstlisting}
\end{devicebox}

\begin{cloudbox}[title=Step 24: cloud agent output (if routed)]
\begin{lstlisting}[style=mytt]
<think>
I have successfully cooled the winebottle 1 using the fridge 1 and now I am at the diningtable 1. The next step is to place the cooled winebottle on the diningtable 1. Since I am already at the diningtable 1, I can directly move the winebottle 1 to the diningtable 1.
</think>
<action>
move winebottle 1 to diningtable 1
</action>
\end{lstlisting}
\end{cloudbox}

\subsection{WebShop}

This section shows a WebShop trajectory coordinated by \name. For routine steps such as query formulation, the device-side execution is sufficient. For decision-intensive steps requiring precise filtering and goal consistency, the cloud becomes decisive: in Step 4, it correctly continues exploration while the device backtracks; in Step 11, it identifies a valid item and completes the purchase, whereas the device fails to recognize task satisfaction. \name effectively learns task-specific behavioral patterns of the device--cloud model pair and makes routing decisions accordingly.

\begin{envbox}[title=Step 1: environment]
\begin{lstlisting}[style=mytt]
You are an expert autonomous agent operating in the WebShop e-commerce environment.
Your task is to: Find me men's polos with short sleeve, button closure with color: black, and price lower than 40.00 dollars.
Your current observation is: "Search".
\end{lstlisting}
\end{envbox}

\begin{herabox}[title=Step 1: Hera decision]
\begin{lstlisting}[style=mytt]
0 (device)
\end{lstlisting}
\end{herabox}

\begin{devicebox}[title=Step 1: device agent output]
\begin{lstlisting}[style=mytt]
<think>
To find men's polos with short sleeves, a button closure, and a black color at a price lower than $40.00, I need to construct a search query that includes these specific criteria. The available actions are "search[<your query>]" and "click[search]". Since the shop is currently observing the search interface, I should use the "search[<your query>]" action to input the query.
The query should include:
- Gender: Men
- Sleeve type: Short
- Closure type: Button
- Color: Black\n- Price range: Less than $40.00
I will formulate the query as follows: "men short sleeve polo button black price <40"
</think>
<action>
search[men short sleeve polo button black price <40]
</action>
\end{lstlisting}
\end{devicebox}

\begin{cloudbox}[title=Step 1: cloud agent output ((if routed)]
\begin{lstlisting}[style=mytt]
<think>
The task is to find men's polos with specific attributes: short sleeves, button closure, black color, and a price lower than $40.00. The current observation is "Search", which means I need to start by searching for the product that matches these criteria. The first step in this process is to input a search query that includes all the required attributes. Since the admissible actions include "search[<your query>]", I should use this action to begin the search.
</think>
<action>
search[men's polos short sleeve button closure black price < 40.00]
</action>
\end{lstlisting}
\end{cloudbox}

\begin{envbox}[title=Step 4: environment]
\begin{lstlisting}[style=mytt]
"Back to Search" [SEP] "Page 3 (Total results: 50)" [SEP] "< Prev" [SEP] "Next >" [SEP] "B085WQKRRJ" [SEP] "CLANMILUMS Men's Classic Comfort Soft Regular Fit Short Sleeve Henley T-Shirt Tee" [SEP] "$17.99 to $19.99" [SEP] "B07N7TDKXQ" [SEP] "SOCKS"NBULK Mens Cotton Crew Neck Short Sleeve T-Shirts Mix Colors Bulk" [SEP] "$38.79 to $172.8"
......
[SEP] "B00O30JLDK" [SEP] "Legendary Whitetails Men's Non-Typical Long Sleeve T-Shirt" [SEP] "$10.52 to $40.5".
\end{lstlisting}
\end{envbox}

\begin{herabox}[title=Step 4: Hera decision]
\begin{lstlisting}[style=mytt]
1 (cloud)
\end{lstlisting}
\end{herabox}

\begin{cloudbox}[title=Step 4: cloud agent output]
\begin{lstlisting}[style=mytt]
<think>
I need to find men's polos with short sleeves, button closure, in black, and priced under $40.00. I have already searched and navigated through the first two pages of results, but I haven't found a suitable product yet. Now, I am on the third page of search results.
Let's review the products listed on this page:
- B085WQKRRJ: CLANMILUMS Men's Classic Comfort Soft Regular Fit Short Sleeve Henley T-Shirt Tee (Price: $17.99 to $19.99) - This is a henley t-shirt, not a polo.
......
None of the products on this page match the criteria for men's polos with short sleeves, button closure, in black, and priced under $40.00. Since I have not found a suitable product yet, I should continue to the next page of results.
</think>
<action>
click[next >]
</action>
\end{lstlisting}
\end{cloudbox}

\begin{devicebox}[title=Step 4: device agent output (if routed)]
\begin{lstlisting}[style=mytt]
<think>
Based on the previous observations, I have navigated through pages 1 and 2 of the search results for men's polos with short sleeves, button closures, and the color black, with prices lower than $40.00. On page 3, there are no results that match the exact criteria of being a men's polo shirt with a short sleeve, button closure, and black color. However, I noticed that some items are close but not exactly what we are looking for.
......
Since we haven't found any more matching items on page 3, let's go back to page 2 and see if there are any additional items that might fit our criteria.
</think>
<action>
click[< prev]
</action>
\end{lstlisting}
\end{devicebox}

\begin{envbox}[title=Step 11: environment]
\begin{lstlisting}[style=mytt]
"Back to Search" [SEP] "< Prev" [SEP] "color" [SEP] "black" [SEP] "dark green" [SEP] "green" [SEP] "light blue" [SEP] "navy blue" [SEP] "orange" [SEP] "white" [SEP] "wine red" [SEP] "yellow" [SEP] "Custom Polo Shirts for Men Design Your Own Personalized Logo Text Photo Polo Shirt Customize Classic Polo Shirt" [SEP] "Price: $5.99" [SEP] "Rating: N.A." [SEP] "Description" [SEP] "Features" [SEP] "Reviews" [SEP] "Buy Now".
\end{lstlisting}
\end{envbox}

\begin{herabox}[title=Step 11: Hera decision]
\begin{lstlisting}[style=mytt]
1 (cloud)
\end{lstlisting}
\end{herabox}

\begin{cloudbox}[title=Step 11: cloud agent output]
\begin{lstlisting}[style=mytt]
<think>
The task is to find a men's polo shirt that meets the following criteria: short sleeve, button closure, black color, and price lower than $40.00. In the previous steps, we have already searched for the product and navigated through the search results. We found a suitable product with the ID B09KXCHWGD, which is a "Custom Polo Shirts for Men Design Your Own Personalized Logo Text Photo Polo Shirt Customize Classic Polo Shirt" priced at $5.99. This product meets all the specified criteria, including the price and color.
In the current observation, we are on the product page of B09KXCHWGD, and we have already selected the color "black." The next logical step would be to complete the purchase by clicking the "Buy Now" button.
</think>
<action>
click[buy now]
</action>
\end{lstlisting}
\end{cloudbox}

\begin{devicebox}[title=Step 11: device agent output (if routed)]
\begin{lstlisting}[style=mytt]
<think>
The goal is to find men's polos with short sleeves, button closure, and a price lower than $40.00. The previous searches have shown some results, but none of them were exactly matching the criteria. The most recent observation indicates that the product "Custom Polo Shirts for Men" is being displayed, which does not match the criteria as it is not specifically a men's polo shirt with short sleeves and button closure.
Given the options, clicking on 'back to search' would allow us to reapply the filters and ensure we get the correct products. This action will help us refine our search and potentially find the desired men's polos.
</think>
<action>
click[back to search]
</action>
\end{lstlisting}
\end{devicebox}


\subsection{AppWorld}

In AppWorld tasks, \name exploits the cloud mainly when stronger code synthesis is needed. While the device can handle API inspection, Step 8 requires composing state checks, branching, API calls, and task termination into an executable loop. The cloud produces this complete control logic, whereas the device only extracts an intermediate variable. This highlights \name’s ability to reserve cloud usage for execution-critical programming steps.

\begin{envbox}[title=Step 1: environment]
\begin{lstlisting}[style=mytt]
I am your supervisor and you are a super intelligent AI Assistant whose job is to achieve my day-to-day tasks completely autonomously.
To do this, you will need to interact with app/s (e.g., spotify, venmo, etc) using their associated APIs on my behalf. For this you will undertake a *multi-step conversation* using a python REPL environment. That is, you will write the python code and the environment will execute it and show you the result, based on which, you will write python code for the next step and so on, until you've achieved the goal. This environment will let you interact with app/s using their associated APIs on my behalf.
Here are three key APIs that you need to know to get more information
# To get a list of apps that are available to you.\nprint(apis.api_docs.show_app_descriptions())
# To get the list of apis under any app listed above, e.g. supervisor\nprint(apis.api_docs.show_api_descriptions(app_name='supervisor'))
# To get the specification of a particular api, e.g. supervisor app's show_account_passwords\nprint(apis.api_docs.show_api_doc(app_name='supervisor', api_name='show_account_passwords'))
Each code execution will produce an output that you can use in subsequent calls. Using these APIs, you can now generate code, that the environment will execute, to solve the task.
My personal email is carbrown@gmail.com and phone number is 7934818191.
Your task is: Keep going to the next song on Spotify until you reach a song I have already downloaded.
Now it's your turn to generate code to solve the task.\nYou should first reason step-by-step about which APIs to call, what arguments to use, and how to build your code block to complete the task.
\end{lstlisting}
\end{envbox}

\begin{herabox}[title=Step 1: Hera decision]
\begin{lstlisting}[style=mytt]
0 (device)
\end{lstlisting}
\end{herabox}

\begin{devicebox}[title=Step 1: device agent output]
\begin{lstlisting}[style=mytt]
<think>
To solve this task, we need to follow these steps:
1. Identify the current song playing on Spotify.
2. Check if the current song has been downloaded.
3. If the current song has not been downloaded, skip to the next song.
4. Repeat steps 1-3 until we find a song that has been downloaded.
First, we need to find out the list of songs available on Spotify. We can use the `show_account_passwords` API from the `supervisor` app to get the Spotify password, then use that to authenticate with the Spotify API.
However, since we are restricted to using only the provided APIs and not external packages like `spotipy`, we will assume that there is a method in the `spotify` app to get the current song details and another method to check if a song has been downloaded.
Let's start by finding out the list of available APIs under the `spotify` app.
</think>
<code>
print(apis.api_docs.show_api_descriptions(app_name='spotify'))
</code>
\end{lstlisting}
\end{devicebox}

\begin{cloudbox}[title=Step 1: cloud agent output (if routed)]
\begin{lstlisting}[style=mytt]
<think>
To solve the task, I need to:
1. Get a list of all my downloaded songs on Spotify.
2. Continuously skip to the next song in the current playback until I find a song that is in the list of downloaded songs.
First, I need to check if there are any APIs available for Spotify that can provide the list of downloaded songs and control the playback. I will start by listing the available apps and then look into the Spotify app's APIs.
</think>
<code>
print(apis.api_docs.show_app_descriptions())
</code>
\end{lstlisting}
\end{cloudbox}

\begin{envbox}[title=Step 8: environment]
\begin{lstlisting}[style=mytt]
Result 7:
{
"song_id": 322,
"title": "Lost in the Labyrinth of Love",
"album_id": null,
"album_title": null,
"duration": 281,
"artists": [
    {
    "id": 36,
    "name":
    "Noah Bennett"
    }
],
"played_seconds": 0,
"is_playing": true,
"is_looping": false
}
\end{lstlisting}
\end{envbox}

\begin{herabox}[title=Step 8: Hera decision]
\begin{lstlisting}[style=mytt]
1 (cloud)
\end{lstlisting}
\end{herabox}

\begin{cloudbox}[title=Step 8: cloud agent output]
\begin{lstlisting}[style=mytt]
<think>
The task is to keep going to the next song on Spotify until a downloaded song is reached. We already have the list of downloaded songs and the current song. Now, we need to check if the current song is in the list of downloaded songs. If it is, we are done. If not, we need to go to the next song and repeat the process.
</think>
<code>
# Check if the current song is in the list of downloaded songs
current_song_id = current_song["song_id"]
is_downloaded = any(song["song_id"] == current_song_id for song in downloaded_songs)
if is_downloaded:
# If the current song is downloaded, complete the task
apis.supervisor.complete_task()
else:
# If the current song is not downloaded, go to the next song
apis.spotify.next_song(access_token=access_token)
# Get the new current song
current_song = apis.spotify.show_current_song(access_token=access_token)
print(current_song)
</code>
\end{lstlisting}
\end{cloudbox}

\begin{devicebox}[title=Step 8: device agent output (if routed)]
\begin{lstlisting}[style=mytt]
<think>
Given the current state, we need to keep moving to the next song in the Spotify queue until we find a song that has been downloaded. We have the list of downloaded songs and the current song.
First, we need to get the song ID of the current song to check against the downloaded songs list.
</think>
<code>
current_song_id = current_song["song_id"]
print(current_song_id)
</code>
\end{lstlisting}
\end{devicebox}

\section{Limitations and Future Work}
\label{appdendix:limit_future}

\paragraph{Limitations}
By studying fine-grained, step-level device--cloud coordination for long-horizon agents, \name improves efficiency and establishes a favorable performance--cost Pareto frontier. However, several limitations remain. First, \name assumes a synchronous, always-connected setting with reliable low-latency communication between device and cloud, and does not account for asynchronous execution or intermittent connectivity, which may limit robustness in mobile or remote environments. Second, \name does not incorporate built-in privacy-preserving mechanisms for offloaded data. Although it reduces the frequency of cloud calls, it does not sanitize transmitted context, which may restrict deployment in privacy-sensitive domains such as healthcare or finance. Third, our study focuses on a binary single-device--single-cloud topology. As a result, \name does not address routing and coordination in more complex multi-agent settings, such as systems with multiple edge devices, specialized services, or hierarchical cloud infrastructure, where coordination overhead can scale non-linearly.

\paragraph{Future Directions.}
To address these limitations, we outline several directions for future research in device--cloud collaboration for LLM agents. Building on \name, we hope to further broaden the applicability, robustness, and scalability of step-level coordination, ultimately enabling more reliable and efficient autonomous systems in real-world environments. Future systems should move beyond the assumption of stable synchronous interaction and explicitly reason about latency, communication uncertainty, and temporary disconnection, allowing adaptive execution under weaker connectivity. It is also important to improve the safety of offloading itself. Incorporating privacy-aware representations, controllable context abstraction, or secure transmission and computation mechanisms would make such frameworks more suitable for domains where sensitive information cannot be directly shared with external providers. More broadly, the current setting opens up opportunities to study coordination at larger system scales. Extending beyond single-device--single-cloud topology toward heterogeneous networks of edge devices, specialized services, and hierarchical cloud resources will require new methods for routing, scheduling, and distributed planning, particularly when workflow dependencies and coordination costs grow rapidly with system size.

\end{document}